\documentclass{book}

\usepackage{morgan}
\usepackage{morchap}
\usepackage{morbeta}
\usepackage[raggedsec]{morplay}
\usepackage{natbib} % Author-Year bibliography
\usepackage[draft]{ttdefs}

\usepackage{imakeidx}
\usepackage{amssymb}
\usepackage{amsmath}

\usepackage{wrapfig}
\usepackage{rotating, graphicx}

\usepackage{verbatim}
\usepackage{algorithm}
\usepackage{algorithmic}
\usepackage[dvipsnames]{xcolor}
\usepackage{makecell}

\usepackage{titlesec}

\titleformat*{\paragraph}{\bfseries}

\newenvironment{packed_item}
{\begin{itemize}
\setlength{\itemsep}{1pt}
\setlength{\parskip}{0pt}
\setlength{\parsep}{0pt}}
{\end{itemize}}

\newcommand\eg{{\it e.g.,~}}
\newcommand\ie{{\it i.e.,~}}

\newcommand\mathv{\mathcal{V}}
\newcommand\mathr{\mathcal{R}}
\newcommand\mathe{\mathcal{E}}

\newcommand\e[1]{{\textit{#1}}}
\newcommand\bd[1]{{\textbf{#1}}}

\newcommand\rel[1]{{\small \fontfamily{phv}\fontshape{it}\selectfont #1}}

\newcommand\word[1]{{{\normalsize \fontfamily{cmtt}\selectfont #1}}}

\newcommand\tripleSRT{$(source,relation,target)$}
\newcommand\tripleverb{$(subject,verb,object)$}
\newcommand\triplepred{$(subject,predicate,object)$}

\newcommand\transp{\top}

\newcommand\isa{\rel{is-a}}

\newcommand\wn{WordNet}

\renewcommand\cite[1]{\citep{#1}}
\newcommand\newcite[1]{\citet{#1}}
\newcommand\posscite[1]{\citeauthor{#1}'s [\citeyear{#1}]}
\newcommand\notcite[1]{}

\definecolor{darkred}{rgb}{.6,.1,.3}
\definecolor{violet}{rgb}{.4,.1,.8}
\definecolor{greyblue}{rgb}{.2,.4,1}

\newcommand{\rescal}{{\scshape Rescal}}

\usepackage[utf8]{inputenc}
\usepackage[T1]{fontenc}

\usepackage{tikz}
\usetikzlibrary{positioning}
\usetikzlibrary{chains}
\usetikzlibrary{scopes}

\usepackage{nohyperref}
\usepackage[pageref]{backref}
\usepackage{url}% for .bib
\renewcommand*{\backref}[1]{}
\renewcommand*{\backrefalt}[4]{%
 \ifcase #1 {\small (Not cited.)}%
 \or {\small (Cited on p.~#2.)}%
 \else {\small (Cited on pp.~#2.)}%
 \fi}

\usepackage{wasysym}

\usepackage{array,ragged2e}

\def\ul#1{$\underline{\smash{\hbox{#1}}}$}
\usepackage[dash,dot]{dashundergaps}
\def\dul#1{$\dashuline{\smash{\hbox{#1}}}$}

\usepackage{fancyhdr}

\makeindex

\counterwithout{footnote}{chapter}

\long\def\symbolfootnote[#1]#2{\begingroup
\def\thefootnote{\fnsymbol{footnote}}\footnote[#1]{#2}\endgroup}

\begin{document}

\frontmatter
%\input{frontmatter} % sample front matter
% INCLUDES PREFACE

\thispagestyle{empty}

\newcommand\titlefiveh[1]{{\centerline{{\huge \bd{{#1}}}}}}
\newcommand\titlefivel[1]{{\centerline{{\Large \bd{{#1}}}}}}

\vspace*{1.0in}
\titlefiveh{\textcolor{blue}{Semantic Relations Between Nominals}}

\vspace*{0.125in}
\titlefivel{\textcolor{blue}{Second Edition}}

\vspace*{0.25in}
\titlefivel{\textcolor{blue}{Vivi Nastase, Stan Szpakowicz,}}
\titlefivel{\textcolor{blue}{Preslav Nakov and Diarmuid \'O S\'eaghdha}}

\vspace*{0.25in}
\titlefivel{\textcolor{blue}{Morgan \& Claypool, 2021 (to appear)}}

\vspace*{0.25in}
\titlefivel{\url{www.morganclaypool.com/toc/hlt/1/1}}

\vspace*{1.0in}
\titlefivel{\textcolor{black}{\e{Chapter 5}}}

\vspace*{0.125in}
\titlefiveh{\textcolor{black}{Semantic Relations and Deep Learning}}

\vspace*{0.25in}
\titlefivel{\textcolor{black}{Vivi Nastase$^*$ and Stan Szpakowicz$^{**}$}}

\vspace*{0.25in}
{\large
\begin{tabular}{c|c}
$^*$ Institute for Natural Language Processing & $^{**}$ School of Electrical Engineering\\
& and Computer Science\\
University of Stuttgart, Germany & University of Ottawa, Canada \\
\texttt{vivi.nastase@ims.uni-stuttgart.de} & \texttt{szpak@eecs.uottawa.ca}
\end{tabular}
}

\vspace*{0.25in}
\titlefivel{\textcolor{black}{by the kind permission of Morgan \& Claypool}}

\mainmatter % sample chapters
\setcounter{chapter}{4}

\chapter{Semantic Relations and Deep Learning}
\label{ch:deep}

% commands only for this chapter
\newcommand\realr{\mathbb{R}}
\newcommand\realc{\mathbb{C}}
\newcommand\gauss{\mathcal{N}}
\newcommand\bball{\mathbb{B}}
\newcommand\emb[1]{{\bf #1}}
\newcommand\embv{\emb{v}}
\newcommand\embr{\emb{r}}
\newcommand\embV{\emb{V}}
\newcommand\norm[1]{\left\lVert#1\right\rVert}

\newcommand\extrafootertext[1]{%
\bgroup
\renewcommand\thefootnote{\fnsymbol{footnote}}%
\renewcommand\thempfootnote{\fnsymbol{mpfootnote}}%
\footnotetext[0]{#1}%
\egroup
}

%% %% section %% %%
\section{The new paradigm}
\label{sec:NNpreamble}

The theoretical foundations of artificial neural networks, inspired by biological processes, were laid in the 1940s \cite{McCulloch1943}. The firing of a neuron would represent a proposition, simulating logical calculus in a (neural) network by the activation or inhibition of connections. The perceptron, the algorithm behind the functioning of a single artificial neuron, was invented in the late 1950s \cite{Rosenblatt1958}. There followed the layered structure of the networks familiar to us now, and the back-propagation mechanism, the core of the learning process in this paradigm. \newcite{Rumelhart1986} showed how the back-propagation mechanism can lead to a useful representation on intermediate hidden layers, when they encoded people and family relationships.

The term \e{deep learning} is rather new, and the ``take-over'' of NLP is quite recent, driven mostly by the advances in hardware that have made the theoretical models computationally feasible and efficient on NLP's large-scale corpora, including the induction of semantic representations\index{semantic representation} of words \cite{Mikolov2013a}. Neural networks were fully formed by the time NLP adopted them. They came with many architectures and with mathematical models which the machine learning community developed over the intervening decades. The interplay goes both ways: the particular requirements of NLP tasks have spurred further developments and innovations.

The adoption of deep learning in work on semantic relations has led to methods and modelling assumptions unlike those explored in the previous chapters. There are differences at several levels.

\paragraph{Modelling.}%
In the work described in the preceding chapters, the process of building representations for relation instances is separate from the model which learns to predict relations. First, relation instances are represented by a specially designed set of features; next, a machine-learning algorithm\index{learning!learning algorithm} works on the training data represented by the chosen formalism. This two-step process is not necessary in the neural framework. Semantic relations and their arguments can be, and often are, encoded (that is to say, modelled) together. The encoding of entities depends on the semantic relations in which they participate, while the encoding of semantic relations depends on the arguments they connect.

\paragraph{Assumptions about relations.}%
Disjointness was one of the desiderata for a ``good'' list of semantic relations. The set of semantic relations would in effect partition the space of relation instances. This constraint was useful in traditional learning, where one seldom allowed an instance to belong to multiple classes. When relations which express world knowledge were added to the mix (\eg \rel{bornIn}, \rel{diedIn}), it became common to have two entities connected by more than one semantic relation. In deep learning, the loss of this constraint is not troublesome. Neural networks can deal quite easily with multi-class learning. This means that one can use richer inventories of semantic relations\index{relation!relation inventory}, such as those coming from knowledge graphs\index{graph!knowledge graph (KG)}, which are often multi-graphs: two vertices can be connected by edges of more than one type.

\paragraph{Data sources.}%
Traditional machine learning usually acts on a collection of instances, represented in a systematic manner. Information from different sources can be combined in one feature vector\index{feature!feature vector}, but the production of feature values for pre-specified features may lead to loss of valuable structural or contextual information. In deep learning, hybrid models easily combine different sources of information such as free-form text and structured knowledge graphs\index{graph!knowledge graph (KG)}. The use of data as a knowledge graph\index{graph!knowledge graph (KG)}---a set of interconnected relation triples---affects the modelling of the arguments and of the relations.

%~\\
\vspace{0.125in}

We begin the chapter with a very high-level overview of deep learning in Section \ref{sec:deepL}. We then revisit the research problems relevant to semantic relations. Deep learning for semantic relations often combines in one architecture the processing of an entire sentence which contains a candidate relation. The matter of representing the meaning of the arguments will be intertwined with the representation of the context\index{context representation} and the relational clues---the expression which connects the relation arguments, and the surrounding text. To make things clearer, and to allow for untried combinations, word representations (attributional features\index{feature!attributional feature}, Section \ref{sec:wordEmbeddings}) are presented separately from relation clues and context (relational features\index{feature!relational feature}, Section \ref{sec:contextDL}). Section \ref{sec:NN_data} discusses concerns around datasets, notably deep-learning solutions to distant supervision\index{learning!distant supervision}: how to get automatically, and handle, large amounts of noisy training data. Section \ref{sec:relationsDL} deals with the learning and modelling of semantic relations, either as particular structures or as neural models; it shows how argument representations and contextual clues are interwoven in various learning models.

The new possibilities in the learning of semantic relations have led to a wide variety of solutions; we survey them here. But new methods crop up even as we write, so this chapter is doomed to remain incomplete. The goal is to give the reader a solid overview of the current topics in relation learning, to elaborate on some of the solutions in the literature, and to point her, whenever possible, toward a reference which presents some of these matters in more detail.

%% %% section %% %%
\section[A High-level View of Deep Learning]{A high-level view of deep learning \newline for semantic relations}
\label{sec:deepL}

Our presentation relies on the reader's exposure to the theory and methods of deep learning. For the uninitiated, there are tutorials and books online. For example, \newcite{Goodfellow2016} give an excellent account of deep learning paradigms and methods.\footnote{\url{www.deeplearningbook.org}.} This section is a brief overview of the main concepts relevant to the task at hand.

A deep-learning algorithm\index{learning!learning algorithm} accepts an input, usually represented as a real-valued vector, and applies to it a function which maps it onto some output values; those values determine the classification decision. In the case of semantic relations, the input represents a relation's arguments, its sentential context, additional relational information from a corpus, or a combination thereof. The output represents the prediction: does the posited relation hold between these arguments? The function and its parameters are the relation model, and it depends on the modelling assumptions and the underlying architecture. This sounds like traditional machine learning but there is an essential difference. In deep learning, model derivation (\ie learning the parameter values) takes multiple steps of back-and-forth processing through the layers of the neural network. As a result, even the input state can be transformed according to the neural architecture, the parameters and the discrepancy between the expected and the computed output.

Consider an example. We can choose to give our neural network information only about a relation's arguments, as a concatenation of the representations of these arguments as real-valued vectors. If there is little training data, the representation can consist of vectors which were pretrained on very large corpora---now commonly known as \e{word embeddings}\index{embedding!word/entity embedding} (see Section \ref{sec:wordEmbeddings}). They can be adjusted during training, or kept fixed. If a large amount of training data is available, the vectors can be seeded with random values which are then adjusted during training, so that in combination with the mapping function, \ie the model, they produce a good approximation of the output, \ie relation labels.

The mapping function can be a scoring function\index{scoring function}. Such a function combines the input vector $\emb{i}$ with the parameters $\emb{r}$ which model a target relation $r$,\footnote{Throughout this chapter, representations---\eg embeddings\index{embedding!word/entity embedding} of entities and relations\index{embedding!relation embedding}\index{relation!relation embedding}---are written in \bd{bold}, and entities and relations in \e{italics}.} with the output as a real value between 0 and 1:
\begin{align}
f(\emb{i},\emb{r}) \in [0,1]
\end{align}
To continue with our example, let us make the following assumptions:
\begin{packed_item}
\item the input \emb{i} consists of the embeddings\index{embedding!word/entity embedding} for the relation's two arguments $\emb{v}_1$ and $\emb{v}_2$, which are real-valued vectors of size $d$: $\embv_1, \embv_2 \in \realr^d$;
\item the relation $r$ is modelled as a $d\times d$ matrix $\emb{r}$: $\embr \in \realr^{d\times d}$;
\item multiplication is chosen to model the interaction between the arguments and the relation.
\end{packed_item}

\noindent In this case, the mapping function will look as follows ($^\transp$ is matrix transposition): 
\begin{align}
f : \realr^d\times \realr^d\times \realr^{d\times d} \rightarrow [0,1]
\end{align}
\begin{align}
f(\emb{v}_1,\emb{v}_2,\emb{r}) = \emb{v}_1^{\transp} \emb{r} ~\emb{v}_2
\end{align}
 
\noindent This is actually a model called \rescal\ \cite{Nickel2011}. $f$ should return 1 if relation $r$ holds between these arguments, 0 otherwise. In practice, the function will return a real value in [0,1].\footnote{A softmax function may be necessary to map the actual value into the [0,1] interval.} During training, the parameters will be adjusted in order to bring the value as close to the actual expected value as possible. During testing, a preset threshold usually helps determine if a new combination $(\embv_i,\embr,\embv_j)$ represents the instance of a relation $r$ which holds between arguments $i$ and $j$ represented by their corresponding vectors. 

A mapping function can be almost arbitrarily complex, and can be implemented by various neural network architectures. Let us engage for a while in name-dropping---and acronym-dropping. At our disposal, there are recurrent neural networks (RNN)\index{recurrent neural network (RNN)}, convolutional neural networks (CNN)\index{convolutional neural network (CNN)} or stacks of different types of neural networks, so as to model different types of interactions between the various parts of the input. Each architecture has its own implementation choices, \eg long short-term memory units (LSTM), rectified linear activation units (ReLU) or gated recurrent units (GRU), each with its own specific properties which make them more suitable for some applications than for others. Long sequences are often encoded with, \eg a bi-directional RNN\index{recurrent neural network (RNN)} using LSTM (BiLSTM\index{LSTM!bidirectional LSTM}). The latest advance is the Transformer architecture, starting with the Bidirectional Encoder Representations from Transformers (BERT)\index{Bidirectional Encoder Representations from Transformers (BERT)} \cite{Devlin2018,devlin-etal-2019-bert}\footnote{\posscite{Devlin2018} truly seminal paper on BERT has started a veritable cottage industry. There are versions named SpanBERT, StructBERT, DistilBERT, BERTje, CamemBERT, FlauBERT, RobBERT, KnowBERT, MobilBERT, BERTweet, RuBERT---with certainly \bd{much} more to come.} and Generative Pretrained Transformer (GPT).\footnote{\url{openai.com/blog/language-unsupervised/}} The input vector itself can be the output of a neural network. 

The parameters of the model are learned during training. The output of a computational unit is a function over the input combined with the unit's weights---its internal parameters. The algorithm uses a \e{loss function} to compare the predicted output of the entire network to the expected output (referred to as the gold standard). The difference between the expected output and the one actually produced, together with a learning rate, determines the amount by which the internal parameters should change so that the error will be reduced. To avoid overfitting the training data, the loss function can include a \e{regularization} factor. This factor biases the model towards a simpler one which obeys specific constraints on the parameters: representations close in the Euclidean space, fewer non-zero weights, and so on.

Dropout \cite{Srivastava2014} can also help avoid overfitting. The idea of dropout is that nodes in the network (both their inputs and outputs) are randomly ignored during training. This, in effect, resembles training in parallel a large number of configurations for a neural network. Such a random configuration of nodes makes the training process noisy, and that forces each of the nodes in a layer to contribute more to the final output, to compensate for the inactive ones. It also simulates sparse activation from a given layer, and that encourages the network to actually learn a sparse representation as a side-effect.

When people look at a sentence with an instance of a semantic relation, they see which parts of the sentence are relevant in deciding if the relation holds. The \e{attention mechanism}\index{attention}---an important enhancement to the neural machinery---allows us to model this insight. An implementation of attention\index{attention} filters the representation of the relation instance through a set of weights, and so boosts the contribution of certain parts of a layer in the network (\eg specific words in the context on the input layer) while limiting the effect of others. These weights, as everything in the model, are learned during training.

This chapter will present several options for each of these aspects of deep learning in the task of semantic relation classification\index{relation!relation classification}. Section \ref{sec:wordEmbeddings} and \ref{sec:contextDL} describe types of input. Section \ref{sec:wordEmbeddings} shows how to represent a relation's arguments given only an unstructured text collection, only a knowledge graph\index{graph!knowledge graph (KG)} or a wordnet, or both. Section \ref{sec:contextDL} shows how to represent the relational features\index{feature!relational feature} when taking into account individual words, word sequences, or phrases with grammatical information. Section \ref{sec:relationsDL} describes architectures which combine the input with internal parameters in a variety of models useful in detecting and classifying semantic relations.

In this chapter, we often write that a representation of words or relations obtained by deep learning is \e{induced}. We want to clarify the term here, because it helps distinguish between what is deliberately learned, and what is a felicitous side-effect. As noted in the foregoing, learning in neural networks means determining iteratively the best values of internal parameters which lead to a good mapping of the input onto the expected output. The process has ``side-effects'', such as the adjustment of the starting input representations, the representation computed by a hidden layer which summarizes the larger-sized input in a useful and compressed manner, and so on. Such side-effects are not the target of the learning process, and their emergence (as it comes about in working with the data provided) can be seen as not deliberate. When a deep-learning formalism discovers semantic relations, the aim is to map an input instance (\eg a sentence) onto a relation type. The representations of arguments, relations or even sentences are a useful by-product.

%% %% section %% %%
\section{Attributional features: word embeddings}
\label{sec:wordEmbeddings}

The identification of the semantic relation for a given pair of arguments relies heavily on a good representation of the meaning of the arguments, and of the context in which they appear---if such context is available. Word embeddings\index{embedding!word/entity embedding} in continuous vector spaces are a type of distributional representation\index{distributional representation}. The vectors are no longer indexed by specific words but by more abstract dimensions, assumed to model some underlying latent semantic characteristics of words or entities. Such a representation projects words/entities/morphemes into a multi-dimensional space, in which distance is a proxy for relatedness or similarity. Depending on the data to be modelled or the task at hand, the source of word embeddings\index{embedding!word/entity embedding} can be distributional information (Section \ref{sec:wefromtexts}), graphs which capture a relational model of meaning (Section \ref{sec:wefromgraphs}), or a combination thereof (Section \ref{sec:wefromtextsandgraphs}).

\subsection{Word embeddings from texts}
\label{sec:wefromtexts}

When a large corpus is available, word meaning can be encoded in a very informative way by distributional representations\index{distributional representation} based on co-occurrences in a window or on grammatical relations. The beauty of such representations is that they are easy to interpret given that the dimensions are themselves words. There are, however, considerable drawbacks.
\begin{packed_item}
\item The vectors are very large: the vocabulary is often on the order of at least $10^5$.
\item The dimensions are words, so they are ambiguous. \eg \e{run} can refer to exercising, standing for office or executing a program.
\item Multiple dimensions can refer to the same thing or perhaps to closely related things, \eg \e{buy} and \e{purchase}.
\item Words as dimensions do not solve the sparseness problem because only the same shared dimension indicates an overlap in meaning between words.
\end{packed_item}

A number of methods have been proposed to induce a representation for words in a space with fewer dimensions, much lower than the dimensions in a distributional representation\index{distributional representation}. The number $d$ of dimensions is a preset parameter. A high value of $d$ will lead to a larger but more precise representation, while a lower value will yield a more abstract representation. Finding the best balance between these options often depends on the task; commonly chosen values for $d$ are in the hundreds.

\begin{figure}[ht]
 \includegraphics[scale=0.75]{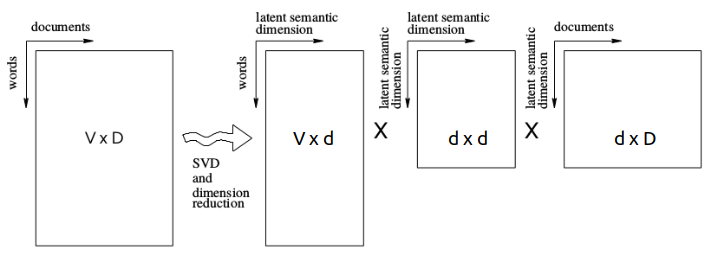}
 \caption{A schematic representation of approximating a word-document matrix by means of Singular Value Decomposition; $V$ is the size of the vocabulary, $D$ is the number of documents in the corpus, and $d$ is the chosen reduced number of dimensions.}
 \label{fig:SVD}
\end{figure}

\newcite{furnas88} applied Singular Value Decomposition (SVD) to the approximation of a word-document co-occurrence matrix. In the process, they uncovered the latent semantic structure of words and documents as low-dimensional vectors in a new space with orthogonal dimensions.\footnote{The dimensions were low in comparison with the size of the vocabulary, which was the base for standard distributional bag-of-words representations.} SVD is presented schematically in Figure \ref{fig:SVD}. This was a step towards projecting words (and documents) into a continuous low-dimensional vector space.

In follow-up work, \newcite{jolliffe02} showed that Principal Component Analysis (PCA), a variation of SVD, can project the word-document vectors into a lower-dimensional space, and that can help reveal the hidden structure of the data.

SVD and PCA rely on a mathematical theory of decomposing a matrix into a product of matrices, each taken to correspond to some part of the input. These methods are applied to fully specified matrices, \ie matrices whose every cell has a defined value.\footnote{This contrasts with adjacency matrices for knowledge graphs---presented later in the chapter---which have mostly unspecified values.} To induce word or document representations using SVD and PCA, one most commonly works with word-document or word-word co-occurrence matrices. The values in such matrices can be either binary (recording simple co-occurrence), or real-valued (getting frequency, perhaps normalized, tf-idf or PMI scores).

Topic modelling seldom has the express purpose of deriving vector representations for words. Even so, topics can be viewed as high-level, abstract, semantic dimensions, and they can be used to produce a representation of words in terms of the probability of their appearance under each of the posited topics \cite{Steyvers2006,Blei2003}.

\newcite{Bengio2003} developed a probabilistic framework for predicting a word from the previously seen words. Every word is encoded as a vector, and a window-based context surrounds a target word. After random initialization, the word representations are adjusted to maximize the probability of the seen text. \citeauthor{Bengio2003}'s innovation was that they used a neural network to encode the probability function of word sequences in terms of the feature vectors\index{feature!feature vector} of the words in the sequence. This allowed the system to learn together the vector representations of the words and the parameters of the function. \newcite{Collobert2008} expanded this framework to multi-task learning.

\begin{wrapfigure}{r}{0.5\textwidth}
\vspace{-2mm}
 \centering
 \includegraphics[scale=0.3]{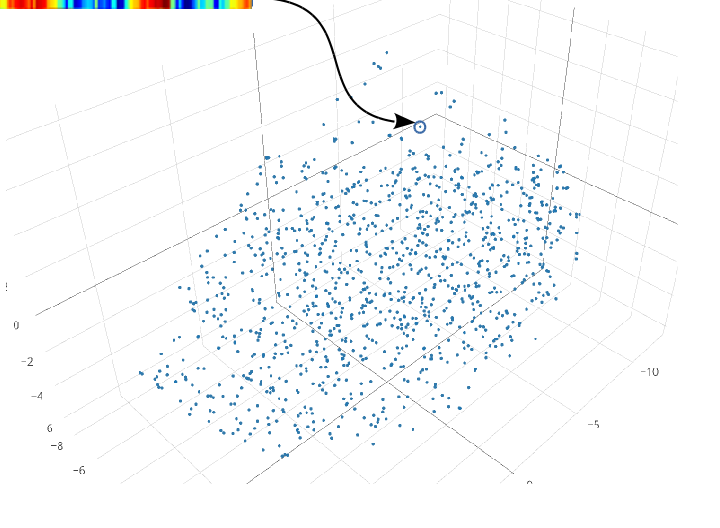}
 \vspace{-3mm}
 \caption{Example of a word embedding\index{embedding!word/entity embedding} as a real-valued vector, projected into a 3D-space for visualization.}
 \vspace{-3mm}
 \label{fig:embedding}
\end{wrapfigure}

Real-valued vectors which represent word meanings have been more widely adopted since deep-learning methods became widespread in NLP, and renamed as \e{word embeddings}\index{embedding!word/entity embedding}; see Figure \ref{fig:embedding}. \newcite{Mikolov2013b,Mikolov2013a} developed two complementary techniques of inducing word embeddings\index{embedding!word/entity embedding}, \ie $d$-dimensional real-valued vector representations of words. The \e{skip-gram model} induces the ``true'' vector for each word by learning to predict the context (the surrounding words) given a word. The \e{continuous bag-of-words (BOW) model} induces word representations while learning to predict a word given its context. 

It has also been noted that the word embeddings\index{embedding!word/entity embedding} induced by this method acquire several types of syntactic and semantic information about words. Such information is reflected as regularities in the relative position of words in the low-dimensional vector space: plurals, derivations, analogies, and so on \cite{Ethayarajh2019}. That allows one to use vector arithmetics on word embeddings\index{embedding!word/entity embedding} as proxies for syntactic and semantic operations on words. 

\begin{wrapfigure}{r}{0.5\textwidth}
\vspace{-7mm}
 \centering
 \includegraphics[scale=0.5]{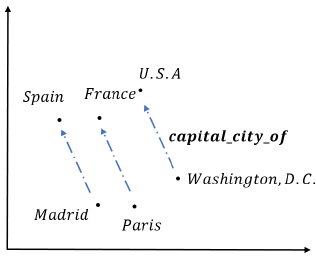}
 \vspace{-3mm}
 \caption{Semantic relations as relative positions of their arguments.}
 \vspace{-5mm}
 \label{fig:relaspos}
\end{wrapfigure}

These operations can also be useful in establishing that different argument pairs are in the same semantic relation. For example, it can be verified that the relative positions of the first and the second arguments are consistent, \eg that the vectors connecting capital cities to their respective countries tend to be parallel; see Figure \ref{fig:relaspos}. Most of the time, there is a more complicated connection between the arguments' position in this space and the relation between them. Nonetheless, even a complex model of relations relies on there being a degree of similarity (maybe only along certain dimensions) between a relation's arguments across numerous instances.

There are many methods of inducing word embeddings\index{embedding!word/entity embedding}. Each method leverages slightly differently the information in a word's context, emphasizes different aspects of a word's meaning, or produces a context-specific embedding. The earliest methods produced ``stand-alone'' embeddings\index{embedding!word/entity embedding}. \newcite{Mikolov2013a} worked with a context window, \newcite{Pennington2014} with grammatical collocations. \newcite{Neelakantan2014,Iacobacci2015} and \newcite{Pilehvar2016} derive word-sense embeddings\index{embedding!word-sense embedding}.

\newcite{Sennrich2016} produce embeddings\index{embedding!word/entity embedding} below the word level. To segment words, they use character n-gram models and a byte-pair encoding compression algorithm. The motivation comes from the problem of out-of-vocabulary words in machine translation.

Word embeddings\index{embedding!word/entity embedding} are, in effect, projections of words into a multi-dimensional space; the words' coordinates in the new space preserve their specific properties. For example, semantically or functionally similar words should be close in such a space. Most models project words onto a Euclidean space, and take various distance metrics as proxies for word similarity. 

\begin{wrapfigure}{r}{0.5\textwidth}
 \centering
 \includegraphics[scale=0.25]{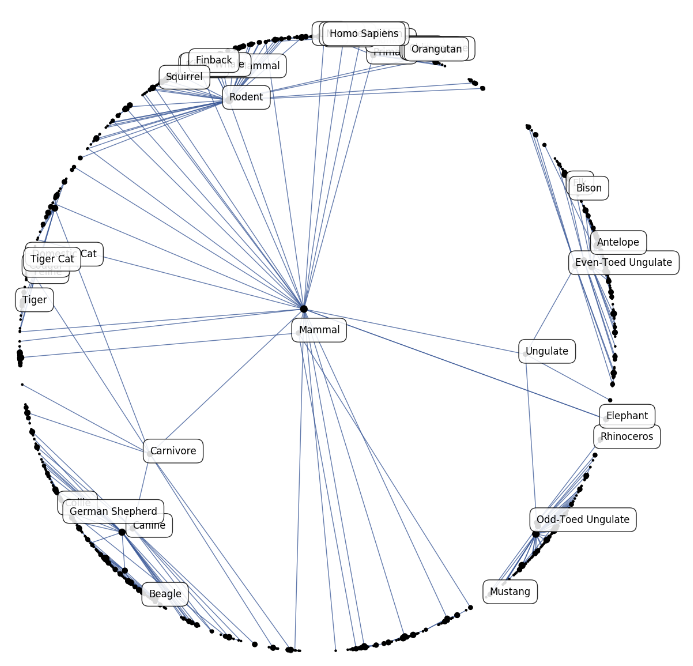}
 \caption{ Two-dimensional Poincar\'{e} embeddings\index{embedding!hyperbolic space embedding} of transitive closure of \wn's\index{WordNet} mammal subtree \cite{Nickel2017}.}
 \label{fig:poincareembs}
\end{wrapfigure}

\noindent
Other types of spaces may be even more appropriate for meaning representation. \newcite{Nickel2017} embed words into a hyperbolic space, or more precisely into a Poincar\'{e} ball; that is a topologically open space bounded by a d-dimensional sphere with unit radius, $B^d = \{ x \in \realr^d~|~\norm{x} < 1 \}$. 

Distances in this space are measured along its equivalent of straight lines: arcs orthogonal to the space's surface boundary. \citeauthor{Nickel2017} show that the properties of hyperbolic spaces make this kind of representation particularly suited to the modelling of hierarchical data. Distances between words in the hyperbolic space mirror distances between nodes in a tree: distances between words closer to the root (situated somewhere close to the center of the ball) will be larger than distances between words closer to the leaves. Figure \ref{fig:poincareembs} shows the projection in two dimensions of the embeddings\index{embedding!hyperbolic space embedding} in the Poincar\'{e} space for words in \wn\index{WordNet}. \newcite{Nickel2018} further optimize this by embedding the taxonomy in the Lorentz model of hyperbolic space.

Contextualized word embeddings\index{embedding!contextualized embedding} are the most recent development. Embeddings from Language Models (ELMo) \cite{Peters2018} employ a neural architecture based on BiLSTMs\index{LSTM!bidirectional LSTM}. ELMo captures different characteristics of the input words at different levels of the neural architecture, corresponding roughly to characters, syntax and semantics; the representation of a word is a combination of the representations at these levels. Bidirectional Encoder Representations from Transformers (BERT)\index{Bidirectional Encoder Representations from Transformers (BERT)} \cite{Devlin2018,devlin-etal-2019-bert} dynamically produce word representations informed by the surrounding words. BERT's architecture---the transformer architecture---is based on attention\index{attention}. Improvements on these models keep coming, for example A Lite BERT (ALBERT) \cite{Lan2020ALBERT} (it separates a ``general'' and lower-dimensional word embeddings\index{embedding!word/entity embedding} from a higher-dimension contextualized representation on the upper levels of the network), Robustly Optimized BERT Pretraining (ROBERTa) \cite{liu2019roberta}, and Text-To-Text Transfer Transformer (T5) \cite{Raffel2019}.

Many of the embedding methods supply pretrained word embeddings\index{embedding!word/entity embedding} in a number of languages, or even cross-lingual data such as XLM \cite{conneau2019unsupervised}. These representations, built from very large corpora, can be used as-is, or they can be fine-tuned (or retrofitted) during the relation-learning process. If the relation dataset is large enough---as are some knowledge graphs\index{graph!knowledge graph (KG)}---the representation of the arguments can be fine-tuned or even learned together with the relation models.

\subsection{Word/entity embeddings from knowledge graphs}
\label{sec:wefromgraphs}

The word embeddings\index{embedding!word/entity embedding} discussed thus far came from distributional representations\index{distributional representation} of words in unstructured texts. Words/entities can also be represented by relational models: their meaning is identified by their relations with other words (as in a wordnet) or other entities (as in a knowledge graph\index{graph!knowledge graph (KG)}). Wordnets and knowledge graphs are symbolic structures but they can be cast into continuous low-dimensional vector spaces. Representations for nodes (words/entities) and edges (relations) can be derived jointly, and these representations encode the relational (graph) structure. Such representations arise from matrix factorization\index{matrix factorization} or can be learned by various types of neural networks.

Matrix factorization\index{matrix factorization} works on the adjacency matrix\index{adjacency matrix} or matrices, representing the graph. The factorization operation has a parallel scoring function\index{scoring function} which combines the representation of entities and relations, and mirrors the factorization split. For example, the bilinear model \rescal\ \cite{Nickel2011} described briefly in Section \ref{sec:deepL} is a matrix factorization\index{matrix factorization} model. The adjacency matrix\index{adjacency matrix} $A_k$ for relation $r_k$ is factorized as $A_k = E^\transp M_k E$; every column of matrix $E$ corresponds to an entity embedding\index{embedding!word/entity embedding}, and $M_k$ is the representation of relation\index{embedding!relation embedding}\index{relation!relation embedding} $r_k$. The scoring function\index{scoring function} parallels this expression; each entry in matrix $A_k$ corresponding to a triple $(e_i, r_k, e_j)$ is computed as
\begin{align}
f_{ijk} = \embv_i^\transp M_k \embv_j
\end{align}

This function has a clear mathematical expression in terms of the representations of the entities and the relation. The scoring function\index{scoring function} can also be learned by a neural network, and then it is modelled by the chosen architecture and its learned parameters. 

Formally, a graph $G = \{\mathv,\mathr,\mathe\}$ is a triple: vertices, relation types and edges.\footnote{There are various terms in graph theory. A \e{vertex} can also be called a \e{node}, and an \e{edge} referred to as an \e{arc}. We will not try to standardize the terminology because the context is quite unambiguous.}
\begin{align}
\begin{array}{l}
 \mathv = \{x_i~|~i=1,n\}\\
 \mathr = \{r_k~|~k=1,m\}\\
 \mathe = \{(x_i,r_k,x_j)~|~x_i,x_j \in \mathv,\ r_k \in \mathr \}
\end{array}
\end{align}

To embed a graph\index{embedding!graph embedding} is to find a representation $\emb{v}_x$ for each vertex $x \in \mathv$, and a representation $\emb{r}_k$ for each relation $r_k \in \mathr$. This is based on information about edges---the relation instances $(x_i,r_k,x_j)$. The usual assumption is that each $\emb{v}_x \in \realr^{d}$ is a $d$-dimensional real-valued vector, with $d$ chosen \e{a priori}. The relation can be a vector, a matrix, a higher-order tensor\index{tensor} (for $n$-ary relations, $n > 2$), and so on.

\begin{figure}[tbh]
 \centering
 \includegraphics[width=\textwidth]{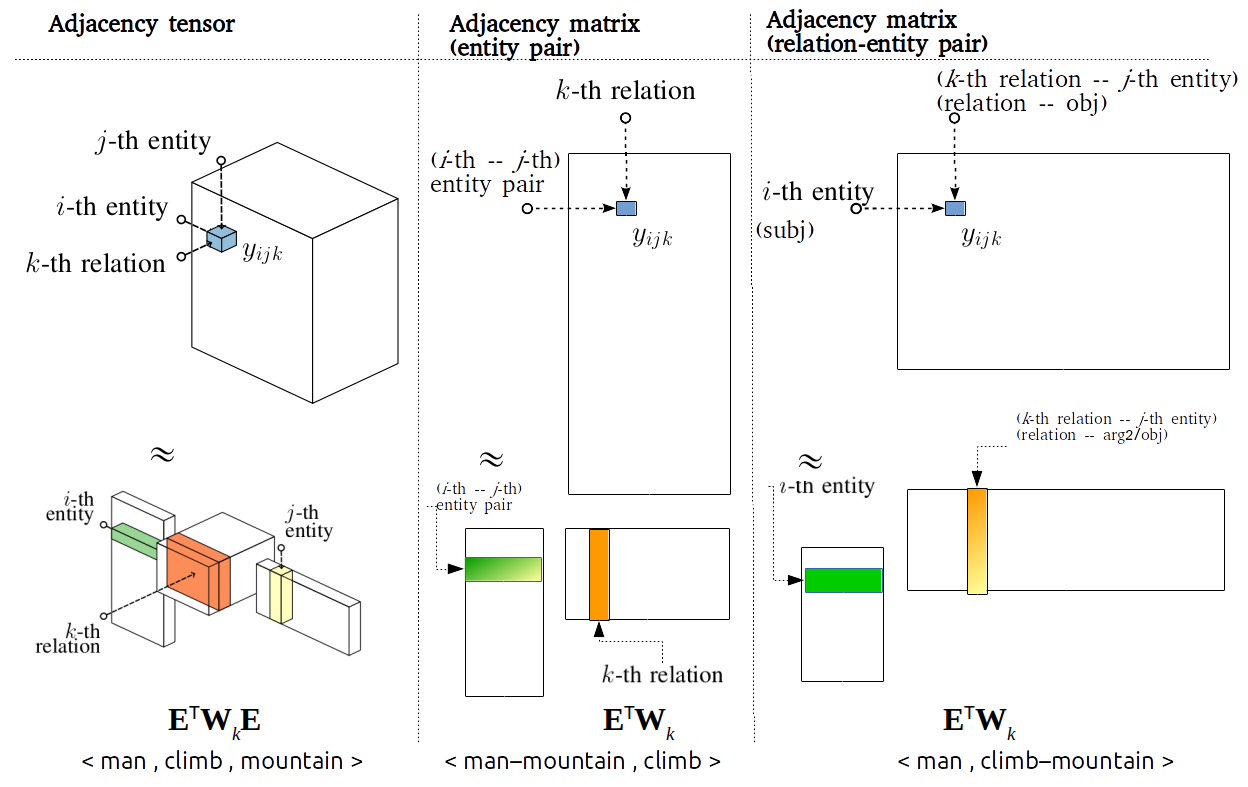}
 \caption{Representations of vertices and relations in a graph for different views of relation tuples. The adjacency tensor\index{tensor}\index{adjacency matrix} illustration comes from \cite{Nickel2016b}.}
 \label{fig:graphrepr}
\end{figure}

\begin{figure}[tbh]
\centering
\includegraphics[width=\textwidth]{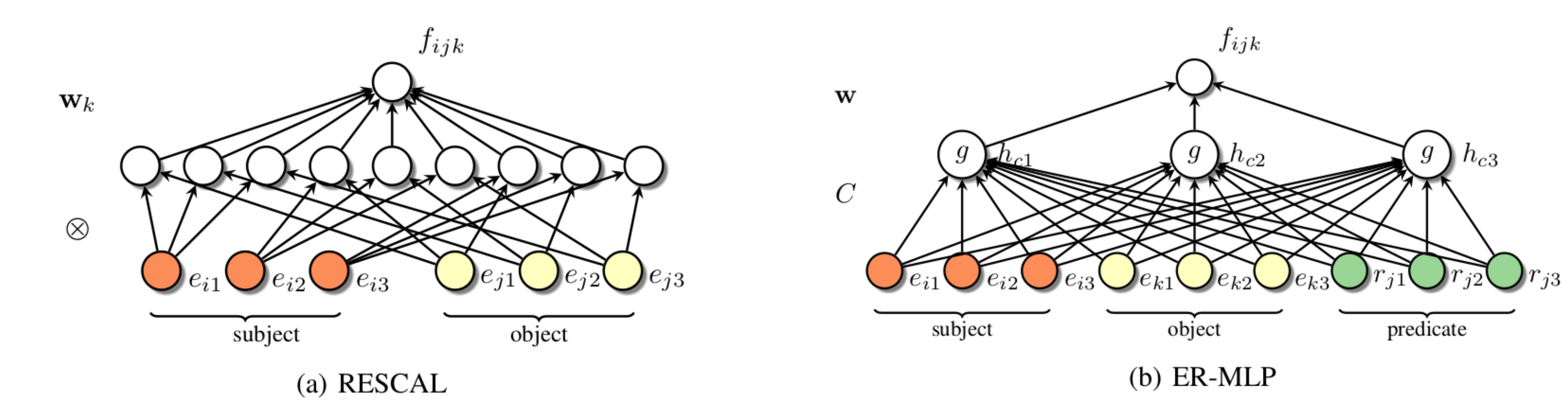}
\caption{Embedding\index{embedding!graph embedding} graphs with neural networks, two examples from \cite{Nickel2016b}. In essence, they learn/implement the scoring function\index{scoring function} $f_{ijk}$. (a) A neural network implements the \rescal\ matrix factorization\index{matrix factorization} model; the representation of the relation is given by the parameters of the hidden layer. (b) A neural network implementation takes a \triplepred\ triple as an input; the embedding of the relation\index{embedding!relation embedding}\index{relation!relation embedding} (\ie the $predicate$) is learned in parallel with the representation of the $subject$ and the $object$.}
\label{fig:nns}
\end{figure}

Other representations are possible. For example, in a knowledge graph\index{graph!knowledge graph (KG)} built from grammatical relations---so that an edge represents a \tripleverb\ triple---it could make sense to represent these as pairs with composite arguments: $(subject\text{--}verb, object)$ or $(subject, verb\text{--}object)$, and so constrain the representation of the arguments with the given relation. Consider an example. If the subject and the verb in $(man, climb, mountain)$ are combined, the pair to be represented will be $(man\text{--}climb, mountain)$. That will constrain the (composite) first argument $man\text{--}climb$ only to objects which a man can climb, as opposed to the more general and separate representations of $man$ and $climb$. Figure \ref{fig:graphrepr} shows the different types of graph representations as adjacency matrices\index{adjacency matrix} obtained when following these various representations of a relation triple.

The adjacency matrix\index{adjacency matrix} or tensor\index{tensor} of a graph, $A$, contains information about the connectivity structure. For a graph containing relation types $r_k$, $A$'s elements are:
\begin{align}
 a_{ijk} = \left\{ \begin{array}{ll}
 1 & if~(i,r_k,j) \in \mathe \\
 NaN & if~(i,r_k,j) \notin \mathe \\
\end{array}\right.
\end{align}

\noindent
1 is the only defined value in $A$ (Not-a-Number is the other). That is because knowledge graphs\index{graph!knowledge graph (KG)} represent only positive instances, \ie only known relation instances. To learn a non-trivial model, some negative instances are required. Section \ref{sec:linkpred} will explain the assumptions needed to produce negative relation instances, as a set $\mathe'$ of ``negative edges''. The corresponding \e{scoring function}\index{scoring function} $f_{i,j,k}$ parallels the information in the adjacency matrix\index{adjacency matrix}, and uses the negative edges to learn non-trivial models:
\begin{align}
f_{ijk} = f(\emb{v}_i, \emb{r}_k, \emb{v}_j) = \left\{ \begin{array}{ll}
 1 & if~(i,r_k,j) \in \mathe \\
 0 & if~(i,r_k,j) \in \mathe' \\
\end{array}\right.
\end{align}

The function can combine the representation of the entities and relations in various ways. Table \ref{tab:graphembs} in Section \ref{sec:linkpred} shows examples. 

Depending on the assumptions about the mathematical form of the representations of entities and relations, and the function $f$, these representations can be induced by matrix factorization\index{matrix factorization}, as illustrated in Figure \ref{fig:graphrepr}, or by deep-learning methods, as shown on two concrete examples in Figure \ref{fig:nns}. \newcite{Levy2014} very nicely explain the equivalences among some of these methods. Comprehensive overviews of such methods appear in \cite{Nickel2016}, \cite{Wang2017} and \cite{Ji2020}. Section \ref{sec:linkpred} discusses the effect of these representation on relation learning.

\begin{figure}[t]
\centering
\includegraphics[scale=0.65]{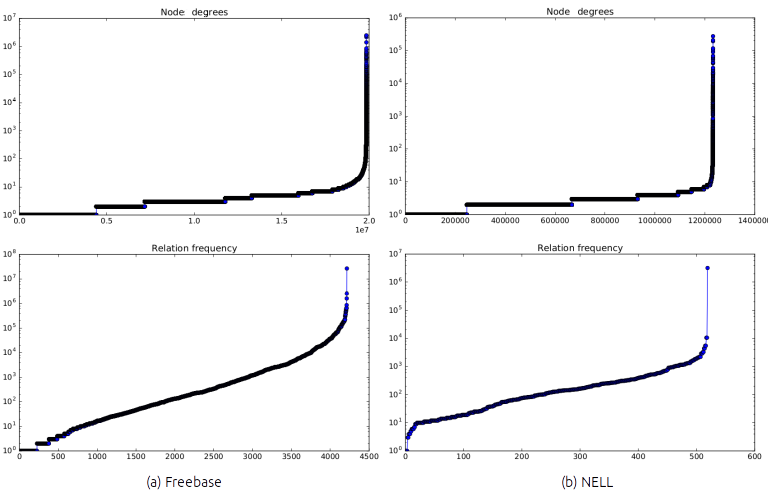}
\caption{Knowledge graph\index{graph!knowledge graph (KG)} statistics on a logarithmic scale: relation and node frequencies for frequently used subsets of Freebase\index{Freebase} and NELL\index{Never-Ending Language Learner (NELL)} (data from \cite{Gardner2014}). Every data point is the degree of a node (top plots), or the frequency of a relation (bottom plots). The data points are ordered monotonically. The $x$ axis is just an index.}
\label{fig:kgstats}
\end{figure}

The structure of the knowledge graph\index{graph!knowledge graph (KG)} directly affects the representations of entities and relations in it. More frequent relations have more informative representations because their adjacency matrices\index{adjacency matrix} are denser. Low-frequency relations, particularly when they connect low-frequency entities, have less informative representations. An analysis of the profile of some of the most frequently used knowledge graphs\index{graph!knowledge graph (KG)} shows that this is a real concern. Figure \ref{fig:kgstats} illustrates it for Freebase\index{Freebase} and NELL\index{Never-Ending Language Learner (NELL)}: most of the nodes appear in very few relations, numerous relations have very few instances. This imbalance can be partially countered by the use of entity type information and relation schemata. Such information can be included as an additional factor in the scoring function\index{scoring function} and in the loss function \cite{Ren2017,Kotnis2017}, or can help organize and optimize adjacency matrix factorization\index{adjacency matrix}\index{matrix factorization} \cite{Chang2014}.

The origins of graph embedding\index{embedding!graph embedding} go back a few decades. \newcite{Rumelhart1986} used a neural network with several hidden layers to learn and predict family relationships. They noted that the network weights and the hidden layers capture representations for the entities (people) and their relationships; that allowed them to predict the second argument of a relation given the first argument and the relation. The weights and node values in the network were not used outside the specific experiment.

\newcite{Paccanaro2002} deliberately set out to induce concept representations from binary relations between concepts in a process they call \e{linear relational embedding}. They aim for $n$-dimensional vector representations of concepts, and $n\times n$ matrix representations of relations. When a relation is $R^c$ applied to a concept $a^c$---multiplying the corresponding matrix $\emb{R}^c$ by the vector $\emb{a}^c$---the result is expected to be a related concept $b^c$ with representation $\emb{b}^c$. \citeauthor{Paccanaro2002} obtain the concept and relation representations by maximizing a \e{discriminative goodness function} $G$. It rewards all concepts which can fill the same $(a^c, R^c, *)$ spot, while maximizing each concept's distance to other concepts nearby (to avoid collapsing all representations to zero).
\begin{align}
G = \sum_{c=1}^C \frac{1}{K_c} log\frac{e^{-\norm{\emb{R}^c \emb{a}^c - \emb{b}^c}_2}}{\sum_{\emb{v}_i\in \embV} e^{-\norm{\emb{R}^c \emb{a}^c - \emb{v}_i}_2}}
\end{align}
\noindent $K_c = |\{(a^c, R^c,*)\}|$ and $\embV$ is the set of all vector representations. This discriminative goodness function is approximated using gradient ascent.

Matrix factorization\index{matrix factorization} for the representation of entities in interconnected data was initially motivated by the goal of clustering\index{cluster, clustering} multi-type interrelated data objects, for example papers, keywords, authors and venues in the domain of scientific publications, or movies, actors and genres in the movie domain \cite{Long2006}. Clustering was achieved by collective factorization of matrices which represent each relation type (\eg \rel{movie\_genre}, \rel{movie\_rating}). Two matrices are related if their row or column indices (\ie their first or second arguments) refer to the same set of objects. \citeauthor{Long2006}'s focus was on clustering, reflected in the matrix factorization\index{matrix factorization} as a product of cluster and cluster association information. The representation of concepts---their association with the induced clusters---is a side-effect not explicitly applied outside these experiments. 

\newcite{Singh2008} address directly the task of relation learning for similar multi-type interrelated data. Unlike \citeauthor{Long2006}, they use matrix factorization\index{matrix factorization} to derive entity and relation representations, and focus on predicting new relation instances in a dataset which covers information about movies (genres, rating, and so on). Each matrix to be factorized represents instances of one relation type. Like in \citeauthor{Long2006}'s work, \citeauthor{Singh2008}'s collective matrix factorization\index{matrix factorization} relies on shared arguments among relations to connect the factors of the different matrices.

The left and right arguments of a relation can have different roles. It may be useful for an entity $a$ to have two different representations, $a_L$ and $a_R$, depending on the role it plays. \newcite{Sutskever2009} induce such a representation by combining topic modelling with matrix factorization\index{matrix factorization}. The latent variables in the topic model represent entity and relation clusters\index{cluster, clustering}. A cluster is represented by its mean and diagonal covariance, and the dual representations for an entity are sampled as vectors from the corresponding cluster. The score of a triple $(a_L,r,b_R)$ is determined by the product of their representations ${\emb{a}_L}^\transp \emb{R} \emb{b}_R$, and that is determined by the clusters to which $a$, $R$ and $b$ belong.

\newcite{Bordes2011} use a neural network to induce entity and relation representations. Entities are represented as $d$-dimensional vectors, and each relation as two $d\times d$ matrices $R_k \approx (\emb{R}^{lhs}_k,\ \emb{R}^{rhs}_k)$. \citeauthor{Bordes2011} hypothesize that if a transformation is applied to each of the two relation arguments $e_i,\ e_j$, then they should become similar. The scoring function\index{scoring function}, then, is this:
\begin{align}
f(e_i,r_k,e_j) = \norm{\emb{R}^{lhs}_k \embv_i - \emb{R}^{rhs}_k \embv_j}
\end{align}

Knowledge graphs\index{graph!knowledge graph (KG)} can also be embedded in non-Euclidean spaces. \newcite{Balazevic2019} introduce a method of embedding a multi-relational graph\index{embedding!graph embedding}\index{embedding!hyperbolic space embedding} in a Poincar\'{e} space. Their Multi-Relation Poincar\'{e} (MuRP) model learns relation-specific parameters which transform the embeddings of source/target entities. If a relation holds between a source and a target entity, their transformed embeddings are close in the Poincar\'{e} space.

\newcite{Weber2018} discuss and compare the types of embedding spaces\index{embedding!embedding space} with respect to their sectional curvature $\kappa$: Euclidean ($\kappa$ = 0), spherical ($\kappa$ = 1), and hyperboloid ($\kappa$ = -1). They note that none of these spaces is optimal for all relational structures. There arises the question of identifying the most suitable embedding space\index{embedding!embedding space} for a given knowledge graph\index{graph!knowledge graph (KG)}\index{embedding!graph embedding}. \citeauthor{Weber2018} analyze the characteristics of local graph neighbourhoods, which they call ``motifs''. A matching space is the space in which the graph motifs can be embedded with the least distortion (or with none). Here are the spaces particularly suited to certain embeddings: hyperbolic spaces (such as the Poincar\'{e} space) for tree structures, spherical spaces for $n$-cycles, and Euclidean spaces for grid structures. The match is apt because the growth rate of these structures matches the growth rate and curvature of the space with respect to its parameters (\eg radius). \citeauthor{Weber2018} compute a heuristic which estimates the growth rate of a graph from its motifs, and map it onto a curvature value. This determines the best space for embedding the graph\index{embedding!graph embedding}.

Nodes in a graph, as well as relations, can also be encoded by methods similar to language models. \newcite{Perozzi2014} transform a (social) network into a set of ``sentences'': sequences of nodes obtained by random walks started on different nodes of the network. Every such sentence represents part of a node's neighbourhood information. They are processed by a method similar to the SkipGram model; the obtained node embeddings\index{embedding!word/entity embedding} maximize the probability that the nodes appear in the observed sequences. \posscite{Wang2020} method is similar; it combines ``sentences''---random walks over the graph---with a multi-layered BiLSTM\index{LSTM!bidirectional LSTM} which encodes contextual information. The entity and relation embeddings\index{embedding!relation embedding}\index{relation!relation embedding} are weighted linear combinations of the model's internal states from each layer of the model.

A graph contains various types of structural information which should be reflected in a node's representation. The neighbourhood of a node can help describe its structural role (\eg as hubs), while dense paths among nodes define node communities. \newcite{Grover2016} aim to develop a graph embedding\index{embedding!graph embedding} method in which the node representations reflect all these structural characteristics. Nodes in a closely connected community should be close in the embedding space\index{embedding!embedding space}. Nodes with the same structural roles should have similar representations, too. The proposed node2vec model finds node representations which maximize the probability of their neighbourhoods. The local neighbourhood of a node is best described by paths found using breadth-first search (BFS), whereas more distant connections and community structure are best captured by depth-first search (DFS). Random walks can produce paths which combine characteristics of BFS and DFS to various degrees. \citeauthor{Grover2016} experiment with two parameters which can bias a random walker towards BFS or DFS. The node representations based on the random walks so obtained lead to state-of-the-art results on a variety of applications, including link prediction\index{link prediction} on Facebook and a protein-protein interaction network. 

\subsection{Word/entity embeddings \newline from texts and knowledge graphs}
\label{sec:wefromtextsandgraphs}

The distributional model\index{distributional representation} and the relational model of language complement each other. Their combination could create a richer and more informative representation of meaning: it would identify both the syntagmatic\index{relation!syntagmatic relation} and the paradigmatic\index{relation!paradigmatic relation} information about words.

There are differences between the unstructured language of texts and the consistent---normalized/canonical---representation of nodes and relations in structured knowledge graphs\index{graph!knowledge graph (KG)}. The differences must be reconciled to take advantage of the information from both these sources. It would be good to be able to ``recognize'', or leverage somehow, concepts from knowledge graphs\index{graph!knowledge graph (KG)} which surface in texts with different lexicalizations; the same goes for relations.

One can add text co-occurrence information---\tripleverb\ triples extracted by an open information extraction\index{open information extraction} system (Open IE)\footnote{\eg \url{stanfordnlp.github.io/CoreNLP/openie.html}}---to the adjacency matrix\index{adjacency matrix}, and factorize this enhanced matrix to learn shared embeddings\index{embedding!word/entity embedding} of entities and relations\index{embedding!relation embedding}\index{relation!relation embedding} in knowledge graphs\index{graph!knowledge graph (KG)} and in text \cite{Riedel2013}. Each argument pair, whether from the graph or from the text, has a corresponding row in the matrix; each relation and predicate has a column. The scoring function\index{scoring function} for modelling the adjacency information in this matrix\index{adjacency matrix} combines a few kinds of data: latent feature compatibility between an argument tuple and a relation, neighbourhood information which benefits from relation similarity, and selectional preferences of relations expressed by the entity models of their arguments.

\citeauthor{Riedel2013} take the atomic view of nodes and relations from the graph, and of arguments and predicates from texts. This means that the procedure relies on (exact) overlaps between arguments and relations in the knowledge graph\index{graph!knowledge graph (KG)} and triples extracted by Open IE. The lexical expressions of predicates and arguments from texts, as well as the forms of relations and nodes from structured knowledge repositories\index{knowledge repository}, can be leveraged to find deeper similarities. For example, the knowledge base\index{knowledge base (KB)} relation \rel{person/organizations\_founded} between a person and the organization which they founded can occur in texts as \word{founder of}, \word{co-founded}, \word{one of the founders of}, \word{helped establish}, and so on. \newcite{Toutanova2015} use a convolutional neural network\index{convolutional neural network (CNN)} to get a vector representation for all textual and knowledge base relations based on their expressions. From these representations, they compute the similarity between predicates and relations, and between nodes and arguments from texts. This similarity is further exploited in the loss function which finds an approximation of the adjacency matrix\index{adjacency matrix} combining knowledge base\index{knowledge base (KB)} information and textual relations. 

\newcite{Alsuhaibani2019} learn hierarchical word embeddings\index{embedding!word/entity embedding} from a corpus and a taxonomy. The taxonomy supplies data about all the ancestors of a word in the taxonomy, and the corpus provides information on the co-occurrence of a word and those ancestors. The signal for tuning the embeddings comes from both sources. The objective function on the taxonomy tries to push the embedding of a word in the taxonomy towards a distance-weighted average of the embeddings of its ancestors. The objective function on the corpus tries to push the embeddings\index{embedding!word/entity embedding} of co-occurring words towards more similarity. 

Section \ref{sec:textgraphDL} gives a deeper overview of the combination of knowledge graphs\index{graph!knowledge graph (KG)} and unstructured texts which aims to derive entity and relation representations for learning semantic relations.

%% %% section %% %%
\section{Relational features: modelling the context}
\label{sec:contextDL}

In the pre-deep-learning work on semantic relations, relational features\index{feature!relational feature} characterize the relation either directly (\eg via an expression or a dependency path\index{dependency!dependency path} between the two relation arguments in a given context), or by background relational features, \ie a collection of patterns\index{pattern} from a large corpus. The length and the syntactic complexity of such expressions vary, so it is problematic to model them formally in order to provide a learning system with a consistent form. Convolution\index{kernel!convolution kernel} and tree kernels\index{kernel!tree kernel} are frequent solutions in traditional machine learning methods. One can also project background relational features\index{feature!relational feature} into a fixed-size low-dimensional space. There are solutions particularly suitable to deep learning: learning a composition function which produces a representation of fixed dimensions---usually a vector, a matrix, or both---for any input string (Section \ref{sec:compositionality} discusses compositionality\index{compositionality}), or representing and using directly a complex tree or graph structure (Section \ref{sec:gnns} discusses graph neural networks\index{graph neural network (GNN)}).

\subsection{Compositionality}
\label{sec:compositionality}

The context---and even the relation arguments---can have variable length. The relevant clues for relation learning can be spread across one or more words, which can appear in different positions and in different grammatical roles. Various methods have been developed to represent such information as a fixed-size data structure and to use it efficiently for relation classification\index{relation!relation classification}. They assemble the representation of a text fragment, taking into account different types of information, \eg the word sequence, the grammatical information, direct dependency paths\index{dependency!dependency path}, or the entire dependency structure. One can also make assumptions about what the final representation will be (a vector, a matrix, or both), and so design an architecture which models an appropriate composition function. This section presents several ways of assembling the textual clues and the context of a relation instance. It focuses on compositionality\index{compositionality} not as a general concept and set of techniques but as specific techniques already tried in the context of relation extraction\index{relation!relation extraction} or classification\index{relation!relation classification}. That is why it will not discuss configurations of recurrent\index{recurrent neural network (RNN)} and recursive neural networks\index{recursive neural network}, convolutional neural networks\index{convolutional neural network (CNN)} or transformers not yet applied in relation learning. In theory, at least, one can use any technique which builds a semantic representation\index{semantic representation} for a text fragment of variable length to represent this kind of contextual information.

\subsubsection{Averaged representation}
\vspace{2mm}

The representation of a phrase as the average of the (distributional) representations\index{distributional representation} of its words is a good approximation of the meaning of a phrase, despite the simplicity and the obvious disregard for word order \cite{Mitchell2010}. A phrase $x$ of $n$ words $x = w_1, w_2, \ldots, w_n$, would be represented as
\begin{align}
\emb{v}_x = \frac{\sum_{i=1}^n \emb{v}_{w_i}}{n}
\end{align}
where $\emb{v}_{w_i}$ is the embedding\index{embedding!word/entity embedding} of word $w_i$. It is trivial to get such a representation for word embeddings\index{embedding!word/entity embedding}, which are real-valued vectors.

Part-of-speech (POS) information and the grammatical role which a word plays in a sentence can add knowledge useful for representing the meaning of a word in context. So, word embeddings\index{embedding!word/entity embedding} can be combined with additional features, \eg syntactic roles or a word's POS, to represent sentence substructures. \newcite{Gormley2015} compute \e{substructure embeddings} $h_{w_i} = f_{w_i} \otimes v_{w_i}$, where $f_{w_i}$ is a vector of hand-crafted features, and $\otimes$ is the outer product. The \e{annotated phrase embedding} sums over the substructure embeddings:
\begin{align}
\emb{v}_x = \sum_{i=1}^n h_{w_i} = \sum_{i=1}^n f_{w_i} \otimes \emb{v}_{w_i}
\end{align}

Such a model can integrate in the low-dimensional continuous representation of words either additional information from those words' (local or global) context, or general information such as types or categories.

\subsubsection{Recurrent Neural Networks}
\label{sec:RNNs}
\vspace{2mm}

Taking into account the word order, a phrase can be encoded as a sequence of words. The representation of a phrase can be derived by a recurrent neural network (RNN)\index{recurrent neural network (RNN)} which combines at each (time) step $t$ the representation of words $w_1, w_2, \ldots, w_{t-1}$ with the current word $w_t$ \cite{Mikolov2010}. An RNN\index{recurrent neural network (RNN)} has an input layer connected to one or more hidden layers, and an output layer. The activation on the hidden layer at the last step is customarily taken as the sequence encoding. The output layer depends on the task (\eg at each step $t$ it is a word in a target language, or the type or topic of the sentence in a classification task), and feedback trains or fine-tunes the input word representations and the weights of the hidden layers. The hidden state is updated at each step $t$ with the representation $\emb{v}_{w_t}$ of the current word $w_t$:
\begin{align}
h_t = f(h_{t-1},\emb{v}_{w_t})
\end{align}
$f$ is a non-linear activation function, \eg an element-wise logistic sigmoid function, or an LSTM/GRU/ReLU unit.

A bi-directional RNN\index{recurrent neural network (RNN)} can also be used. At each time step $t$, the hidden layer combines two representations: one for the forward sequence (as for the regular RNN), the other for the backward sequence (the input phrase in reverse order, to allow the model to see the ``future'', \ie the upcoming words).

A recurrent neural network\index{recurrent neural network (RNN)} helps ensure that word order is accounted for, and that a word sequence of arbitrary length can be encoded as a fixed-length vector which then serves as an input to a relation classifier\index{relation!relation classification}, typically another neural network. In practice, RNN\index{recurrent neural network (RNN)} units such as LSTMs \cite{LSTMs} or GRUs \cite{GRUs} are used to address such matters as vanishing (or exploding) gradients during back-propagation.\footnote{\url{www.cs.toronto.edu/~rgrosse/courses/csc321_2017/readings/L15\%20Exploding\%20and\%20Vanishing\%20Gradients.pdf}}

\subsubsection{Recursive Neural Networks}
\vspace{2mm}

\begin{figure}[tbh]
 \centering
 \includegraphics[scale=0.3]{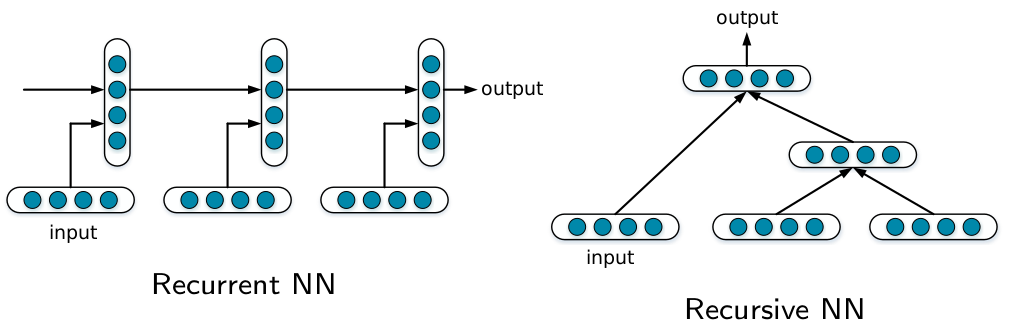}
 \caption{Recurrent\index{recurrent neural network (RNN)} and recursive neural networks\index{recursive neural network} \cite{Socher2011:parsing}.}
 \label{fig:rnns_2}
\end{figure}

While RNNs\index{recurrent neural network (RNN)} model word order, word relations beyond linear order, \eg grammatical structure, might also be worth modelling. The next level of complexity are recursive neural networks\index{recursive neural network}, which can create a bottom-up representation for a tree-structured context by recursively combining representations of sibling nodes---see Figure \ref{fig:rnns_2}.

Given a phrase $x$ of $n$ words, $x = w_1, \ldots, w_n$, and a tree which represents its syntactic structure in some formalism, a recursive neural network\index{recursive neural network} assembles the representation of the phrase bottom-up:
\begin{align}
\emb{a}_{i,j} = f(\emb{a}_i,\emb{a}_j)
\end{align}
$\emb{a}_{i,j}$ is the representation of the node $a_{i,j}$ in the hierarchical structure of the phrase, with children $a_i$ and $a_j$. A child can be an internal node in this structure, assembled from the representation on \e{its} children, or a leaf node. For the latter, the representation will be the embedding\index{embedding!word/entity embedding} of the corresponding word: $\emb{a}_i = \emb{v}_{w_i}$. The function $f$ can take different forms, just as it does for RNNs\index{recurrent neural network (RNN)}. 

\subsubsection{Incorporating dependency paths}
\vspace{2mm}

The methods noted before---the averaged representation and the composition via RNNs\index{recurrent neural network (RNN)} and recursive\index{recursive neural network} neural networks---have slowly incorporated more and more of the available contextual information, including grammatical structural information. The next step is to include information about the type of grammatical relations which connect the nodes in the tree or graph representation of the context\index{context representation} of a relation instance.

Relational features\index{feature!relational feature} in earlier work on semantic relations pick out evidence about two entities' interaction in a given context. One of the successfully applied types of relational features is the dependency path\index{dependency!dependency path} which connects the potential relation arguments.

\begin{figure}[tbh]
\centering
\includegraphics[clip=true, trim=0cm 1cm 0cm 0cm, scale=0.4]{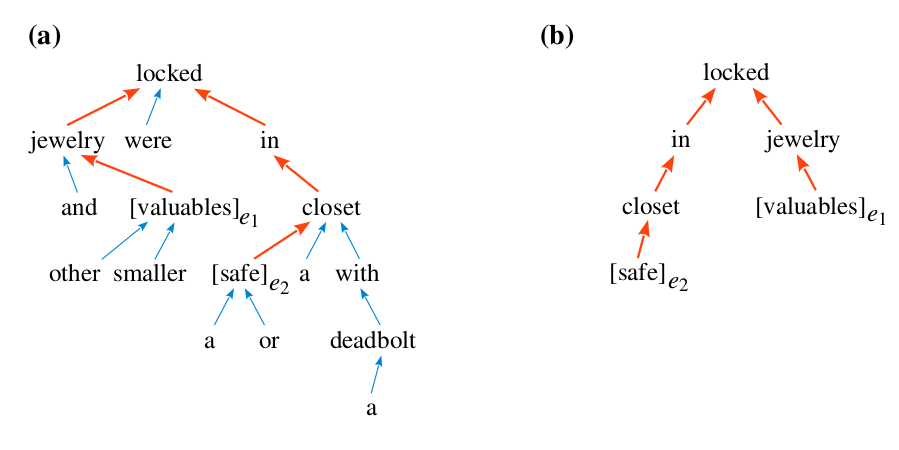}
\caption{The dependency path\index{dependency!dependency path} (red) between entities $e1$ and $e2$ in the sentence ``Jewelry and other small [valuables]$_{e1}$ were locked in a [safe]$_{e2}$ or a closet with a deadbolt.'' \cite{Xu2016}.}
\label{fig:xu2016deps}
\end{figure}
\begin{figure}[tbh]
\centering
\includegraphics[clip=true, trim=0cm 1.3cm 0cm 0cm, scale=0.35]{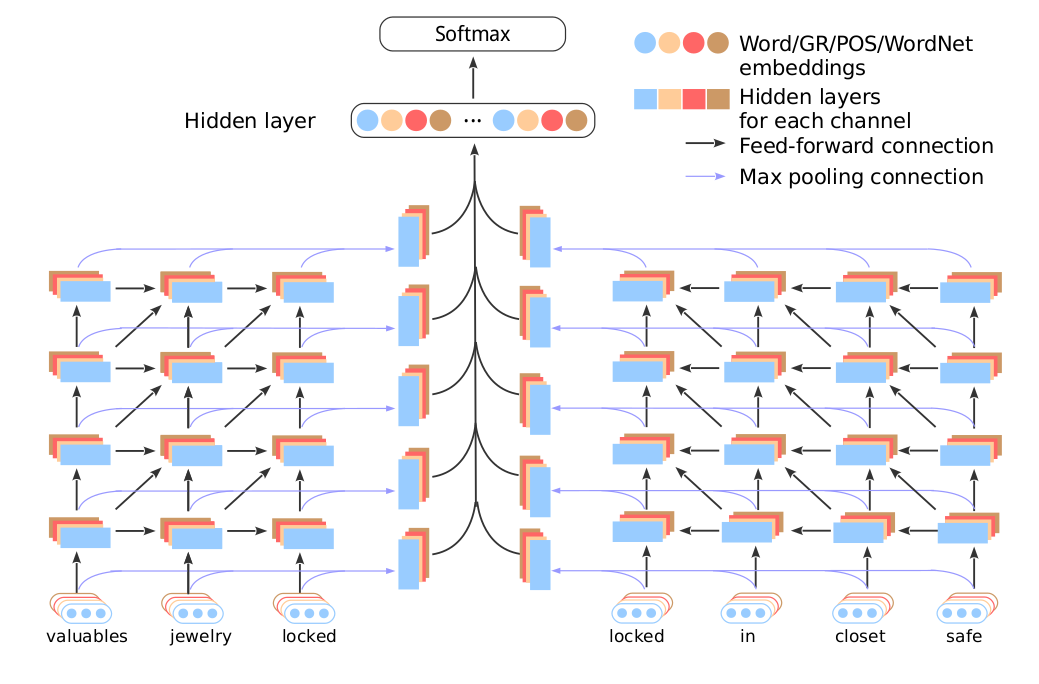}
\caption{The two parts of the dependency path\index{dependency!dependency path}---separated by the common ancestor---are encoded by RNNs\index{recurrent neural network (RNN)} over four information channels: words, part-of-speech tags, grammatical relations, and WordNet\index{WordNet} hypernyms\index{relation!hypernymy} \cite{Xu2016}. Relation prediction is based on a final hidden layer, in which these representations are combined.}
\label{fig:xu2016rnn}
\end{figure} 

When it comes to the dependency structure, several levels of information can describe the connection between two words in a sentence. The first level is the dependency path\index{dependency!dependency path}, a linear chain or a tree with two linear branches. The nodes on this basic path can have more dependencies, which lie outside the path of interest but may add information relevant to their meaning or role in the path. Such ``side'' dependencies makes it an \e{augmented dependency path}\index{dependency!dependency path!augmented dependency path} \cite{Liu2015}, with a more complex tree structure which can be encoded with string/tree/graph kernels. In deep learning, it can also be encoded with various types of neural networks which gradually assemble the context into a fixed-size input, and in effect implement a compositionality\index{compositionality} function.

The dependency path can be viewed as two branches which join the relation arguments with a common ancestor \cite{Xu2016}. Either branch can be encoded separately, and with various types of information: words, parts of speech, grammatical relations, WordNet\index{WordNet} hypernyms\index{relation!hypernymy}, and so on. The dependency relations\index{dependency!dependency relation} depicted in Figure \ref{fig:xu2016deps} are encoded by means of deep RNNs\index{recurrent neural network (RNN)}, as shown in Figure \ref{fig:xu2016rnn}. Relation prediction is based on a vector representation which combines the outputs of the encoding of this multi-layered information of the dependency paths\index{dependency!dependency path}.

Each word in a dependency path\index{dependency!dependency path} may have additional relations which clarify its semantics and its role in the phrase. Some of this information may help recognize the semantic relation between the target arguments. \newcite{Can2019} use the richer-but-smarter shortest dependency path\index{dependency!dependency path}---augmented with dependent nodes\index{dependency!dependency path!augmented dependency path} selected by various attention\index{attention} mechanisms with kernel filters. This smartly augmented path is then processed by a CNN\index{convolutional neural network (CNN)}.

\subsubsection{Compositionality models}
\vspace{2mm}

\begin{figure}[tbh]
\centering
\includegraphics[scale=0.4]{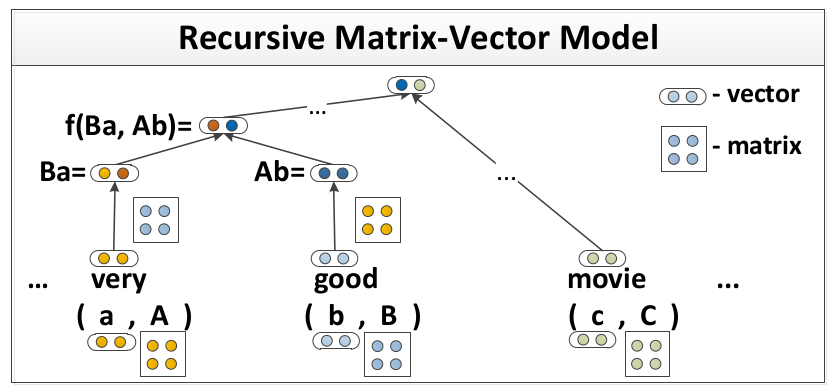}
\caption{A recursive neural network\index{recursive neural network} which learns semantic vector representations of phrases in a tree structure. Each word and each phrase is represented by a vector and a matrix, \eg \e{very = (a, A)}. The construction of the representation of a phrase, \eg \e{very good}, is based on the representations of its words \cite{Socher2012}.}
\label{fig:compositionality}
\end{figure}

The methods just discussed approach compositionality\index{compositionality} gradually, by combining semantic representations\index{semantic representation} of words. This ranges from simple averaging to the use of grammatical structure in assembling the meaning of a phrase. Grammatical relations, while also used, are not modelled explicitly. This next step models the grammatical relations themselves, either as part of word semantics or separately.

A word's embedding\index{embedding!word/entity embedding} is induced from the contexts in which it appears. This represents a variety of aspects related to the word's form and meaning, as properties of its position in the embedding space\index{embedding!embedding space} relative to the position of its morphologically inflected forms, or other words \cite{Mikolov2013a,Levy2014lingreg,Finley2017,Ethayarajh2019}. It may be desirable to build word representations which address specific aspects relevant to assembling the representation of a phrase. In particular, they can have separate components to model the meaning of a word and its ``composition\index{compositionality} function'', essentially an operator which encodes how the word modifies the meaning of another word it combines with. The meaning vector and the composition\index{compositionality} function of a phrase can be recursively assembled from their constituents in \e{recursive matrix-vector spaces} \cite{Socher2012}. The word's meaning is modelled as a vector, and its composition\index{compositionality} function as a matrix, as shown in Figure \ref{fig:compositionality}.

These representations are induced during training. Words whose semantic component is stronger (\eg content words such as nouns or verbs) will have a more informative semantic component. Words with a more structural role (\eg function words such as prepositions or conjunctions) will have a more informative compositional\index{compositionality} component. Both components would probably be equally strong for any content-altering modifier (such as \e{fake}) or for a verb which functions as a hub for the event it signals.

\begin{figure}[tbh]
 \centering
 \includegraphics[scale=0.6]{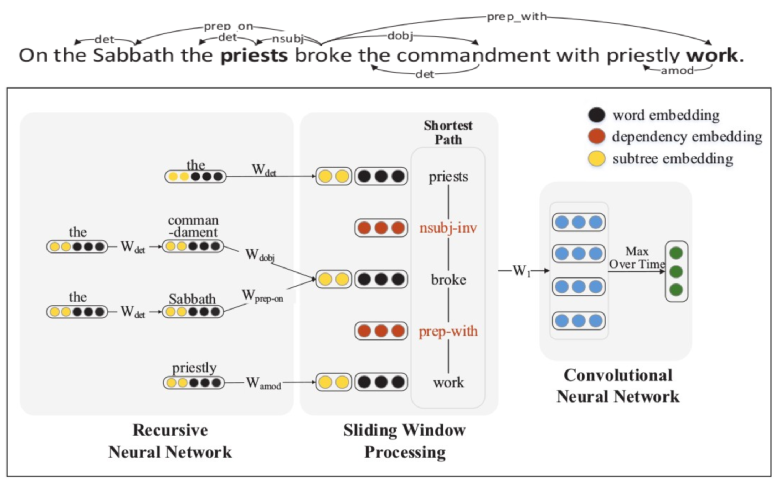}
 \caption{Recursive representation of a phrase based on the augmented dependency path\index{dependency!dependency path!augmented dependency path} and on the dependency relations\index{dependency!dependency relation} \cite{Liu2015}.}
 \label{fig:liu2015rec}
 \end{figure}

Dependency relations\index{dependency!dependency relation} can also be encoded explicitly. This keeps the composition information outside a word's representation, and allows different combinations to take syntactic information into account. \newcite{Liu2015} encode augmented dependency paths\index{dependency!dependency path!augmented dependency path}. They use vector encodings of dependency relations\index{dependency!dependency relation} and dual representations for words: a word's semantics, and the subtree it dominates. The (shortest) dependency path\index{dependency!dependency path} between two entities is augmented with the subtrees dominated by each head word along the path. The network recursively assembles the representation of a phrase which connects two entities, using the augmented dependency path\index{dependency!dependency path!augmented dependency path} and the dependency relation\index{dependency!dependency relation} representations. Figure \ref{fig:liu2015rec} illustrates.

\subsubsection{Transformer-based sentence embeddings}
\vspace{2mm}

The methods we have reviewed thus far assemble the representation of a text fragment gradually from pre-trained or learned word representations. The representation of a word is fixed, regardless of the context in which the word appears. Transformer-based methods tackle the problem differently: a text fragment (often a sentence) is directly encoded, and a word may have different representations for different contexts. An adaptation of transformers for relation learning poses the problem of providing information on the relation's arguments. Such information enables the system to learn the targeted relation\index{relation!targeted relation}, and to assemble the fixed-length relation representation from the various layers of information in the transformer.

\newcite{Soares2019} change a transformer into such a relation encoder. They experiment with various ways of supplying information about the location of the arguments in the input text, and with different learning set-ups. The first set-up was relation classification\index{relation!relation classification} from manually annotated data. The results were the best when entity markers (special tokens [{\small E1}$_{start}$], [{\small E1}$_{end}$], [{\small E2}$_{start}$], [{\small E2}$_{end}$]) signaled the start and end positions of the two relation arguments in the text fragment, and when the concatenation of the final hidden states corresponding to [{\small E1}$_{start}$] and [{\small E2}$_{start}$] was taken as a relation instance representation. This mirrors the use of the output state which corresponds to the special [{\small CLS}] token as the sentence representation\index{Bidirectional Encoder Representations from Transformers (BERT)} \cite{Devlin2018,devlin-etal-2019-bert}.\footnote{The first token of every sequence in BERT is a special classification token [CLS].}

In the second set-up, distant supervision\index{learning!distant supervision}, \citeauthor{Soares2019} take the transformer configuration developed for the classification task, and aim to produce and then compare the relation representations for pairs of entities in context. The loss function in this case is adjusted to lead to similar representations for relations which link the same pairs of entities. To encourage the system to incorporate contextual information and avoid excessive reliance on the entities in a pair, a special [{\small BLANK}] token replaces one or both of them in the automatically annotated corpus. For each positive instance, \citeauthor{Soares2019} sample negative examples which do not contain the same entity pair, and use contrastive estimation to learn to rank positive instances higher than those presumed negative.
 
\subsection{Graph Neural Networks for encoding syntactic graphs}
\label{sec:gnns}

The foregoing was a survey of the mapping of a phrase relevant to relation classification\index{relation!relation classification} onto a fixed-sized representation, which can be used as input to a neural network for relation classification. The success of such mappings depends on the method of composing the meaning of the larger phrase from its atomic components and from structural information. The encoding of a phrase can also be based on its constituency or dependency graph structure. Previous neural architectures which expect a sequence as input require preprocessing to linearize the graph. This is troublesome: a graph does not have one natural order, unless it is a linear chain. The output of a model which encodes a graph should not depend on the input order of the nodes. Since the patterns possibly relevant to relation learning may in fact be structural patterns\index{pattern} in the graph, it is advantageous to encode the graph structure rather than one of its linear projections.

The recursive neural networks\index{recursive neural network} discussed in the preceding section do encode a non-linear structure: the directed acyclic graph (DAG)\index{graph!directed acyclic graph (DAG)}. They still require preprocessing of the input to decide how this information is to be presented to the neural network. They also can only process certain types of graph structures.

\newcite{Scarselli2009} introduced a connectionist model, \e{graph neural networks}\index{graph neural network (GNN)}, which subsumes recursive neural networks\index{recursive neural network}. A GNN models the structure of a graph via functions which aggregate a node's local or even wider neighbourhood, and it iteratively updates an initial graph representation. To learn the representation of the graph, the GNN minimizes a loss function which captures the difference between the task-dependent predicted output and the gold standard. \citeauthor{Scarselli2009}'s GNN\index{graph neural network (GNN)} model works on homogeneous undirected graphs. Further work has produced models for directed, heterogeneous, dynamic and other types of graphs; there is an overview in \cite{Zhou2018}. 

Our discussion here focuses on a few models which have been applied to encoding the structured textual context\index{context representation} for relation instances, in particular on dependency graphs\index{graph!dependency graph}\index{dependency!dependency graph}. (Section \ref{sec:gnnskg}, in the segment of the book devoted to relation learning, will look at the encoding of large knowledge graphs\index{graph!knowledge graph (KG)} using GNNs\index{graph neural network (GNN)}.) From the point of view of structured textual context, of particular interest are the aggregation functions which encode a node's neighbourhood, and the update steps. The aggregation functions, apart from capturing the neighbourhood structure of the nodes, can incorporate additional information, such as attributes of the nodes and of the edges, \eg bags of words, geolocation, timestamps, images.

The dependency path between the arguments of a relation can be regarded as a tree rooted in a common ancestor. From the standpoint of deep learning, such a structure can be encoded as a bidirectional (top-down and bottom-up) tree-structured LSTM-RNN\index{LSTM!tree-structured LSTM-RNN} \cite{Miwa2016}. The bidirectional model ensures that the information from the root of the path and from the leaves is propagated to each node. Weight matrices for same-type children are shared, and they allow for a variable number of children. This model can encode either the full dependency tree\index{parsing!dependency tree}\index{dependency!dependency tree}, or the shortest path between the relation arguments and the sub-tree, \ie the tree below the lowest common ancestor of the target nodes.

As noted earlier, the augmented dependency path\index{dependency!dependency path!augmented dependency path} includes dependency information on the words on the path. Too much of such information can distract from the relevant portions. One way to control the path is to prune the augmented dependency path\index{dependency!dependency path!augmented dependency path}: the syntactic tree is pruned below the lowest common ancestor by removing tokens further than $k$ steps away from the dependency path\index{dependency!dependency path} between the target words. This ensures that negation and relevant modifiers are kept, while the size of the tree is reduced \cite{Zhang2018}.

\begin{figure}[t]
\includegraphics[scale=0.33]{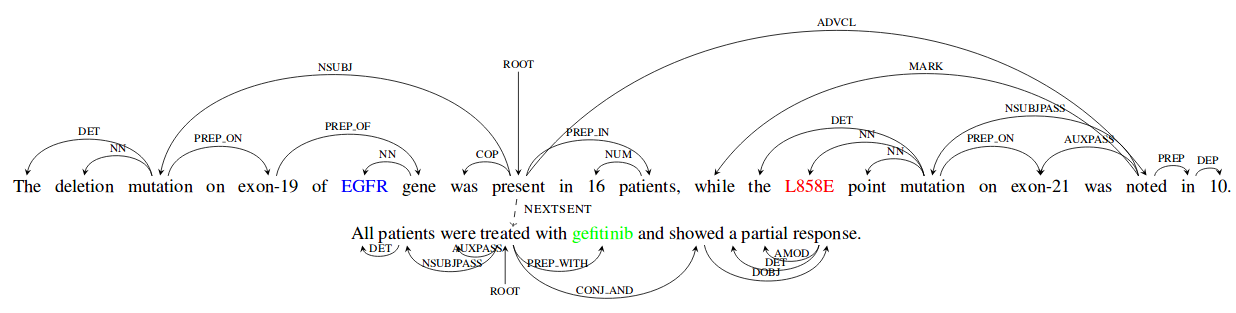}
\caption{Example of a document relation graph, obtained using sequential and syntactic relations; discourse relations are omitted for clarity \cite{Peng2017}.}
\label{fig:docgraph}
\end{figure}

Graph neural networks\index{graph neural network (GNN)} can also help tackle cross-sentence relations. Links beyond the sentence level can be established by sequential or discourse relations\index{relation!discourse relation}. Figure \ref{fig:docgraph} shows a document representation which incorporates intra- and inter-sentential dependencies\index{dependency}, such as sequential, syntactic and discourse relations\index{relation!discourse relation}.

This structure can be encoded with graph LSTMs\index{LSTM!graph LSTM} by partitioning the document graph into two directed acyclic graphs (DAGs)\index{graph!directed acyclic graph (DAG)}. One DAG contains the left-to-right linear chain and other forward-pointing dependencies\index{dependency}. The other DAG covers the right-to-left linear chain and the backward-pointing dependencies\index{dependency}. The effect is a mapping of the graph structure into a BiLSTM\index{LSTM!bidirectional LSTM} formalism \cite{Peng2017}. This representation is used to learn a contextual representation\index{context representation} for each word. Such representations give the input to a relation classifier\index{relation!relation classification} either by simple concatenation, if the arguments are single words, or by first building an averaged representation for multi-word terms.

%% %% section %% %%
\section{Data}
\label{sec:NN_data}

Neural networks are powerful but they require copious training data because they must learn many parameters. Some datasets constructed earlier have been used in deep learning, although mostly as test data because of their small size. Section \ref{sec:kbs} reviews additional datasets created and applied in this framework. Just as in traditional learning, distant supervision\index{learning!distant supervision} methods have been developed, taking advantage of particular characteristics of deep learning to deal with automatically annotated noisy data. Section \ref{sec:distsupnn} describes a few of the deep-learning methods of coping with noisy data, such as adversarial networks\index{generative adversarial network (GAN)}\index{learning!adversarial learning} and reinforcement learning\index{learning!reinforcement learning}.

\subsection{Datasets}
\label{sec:kbs}

Wikipedia infoboxes\index{Wikipedia!Wikipedia infobox} were one of the sources of clean relation instances needed in relation extraction\index{relation!relation extraction}. This was the starting point of Freebase\index{Freebase}, a collaboratively built database, currently available from Wikidata\index{Wikidata}.\footnote{www.wikidata.org/} Various subsets of these data have been in use, frequently for link prediction\index{link prediction} methods, \eg \cite{Socher2011, Trouillon2017,Gardner2015}, or for distant supervision\index{learning!distant supervision} based on knowledge graphs\index{graph!knowledge graph (KG)}. Other knowledge graphs, taken from such resources such as NELL\index{Never-Ending Language Learner (NELL)} and \wn\index{WordNet}, have also played a role in link prediction\index{link prediction} and as sources for distant supervision\index{learning!distant supervision}. Table \ref{tab:dataset} shows the statistics of some of the datasets most commonly used in relation learning.

\begin{table}[ht]
\caption{Knowledge graph\index{graph!knowledge graph (KG)} datasets used for link prediction\index{link prediction} or as sources for distant supervision\index{learning!distant supervision}. {\small (The question marks signal the absence of published statistics. 79.5\% of TACRED instances are ``no relation''.)}}
\label{tab:dataset}
\plaincenterline{\mytable
\begin{tabular}{l|ccr}
\hline
\cb Data set & \cb \# entities & \cb \# relation instances & \cb \# relation types \\
\hline
FB & 20M & 67M & 4,215 \\
\cy FB15K & \cy 14,951 & \cy 600k & \cy 1,345 \\ 
FB \cite{Mintz:2009} & 940k & 1.8M & 102 \\
\cy FB \cite{Riedel2010} & \cy ? & \cy 743k & \cy 53 \\
\hline
NELL & 1.2M & 3.4M & 520 \\
\hline
\cy WN18 & \cy 40,943 & \cy 150k & \cy 18 \\
\hline
Google RE & ? & 54k & 5 \\
\cy GDS & \cy ? & \cy 18,824 & \cy 5 \\
\hline 
FewRel & ? & 70,000 & 100 \\
\cy FewRel 2.0 & \cy ? & \cy 72,500 & \cy 125 \\ 
\hline
TACRED & ? & 106,264 & 42 \\ 
\hline
\cy DocRED & \cy 132,375 & \cy 63,427 & \cy 96 \\
DocRED (DS) & 2,558,350 & 1,508,320 & 96 \\ \hline
\end{tabular}
\emytable}
\end{table}

\newcite{Sun2013} discuss the Google Relation Extraction\index{relation!relation extraction} (RE) corpus.\footnote{\url{code.google.com/p/relation-extraction-corpus/}} It consists of instances of five binary relations: \rel{perGraduatedFromInstitution}, \rel{perHasDegree}, \rel{perPlaceOfBirth}, \rel{perPlaceOfDeath}, and \rel{NA} (none of the above). The corresponding sentences come from Wikipedia\index{Wikipedia}. Annotation was manual but the instances do contain noise, also in the test partition. \newcite{Jat2018} introduced a variation of this dataset, the Google Distant Supervision\index{learning!distant supervision} dataset. It starts with the Google RE relation triples, and adds sentences which contain the relation's arguments in the triple, found by searching the Web.

\newcite{Zhang2017} present TACRED.\footnote{\url{nlp.stanford.edu/projects/tacred/}} This dataset consists of sentences extracted from data created for the TAC Knowledge Base Population\index{knowledge base (KB)!knowledge base population (KBP)} tasks. Sentences containing one of 100 target entities were extracted. The annotators were asked to mark subject and object entity spans\index{entity mention span}, and the relation between them. The dataset, split into training, development and test subsets, contains both positive and negative instances.

\posscite{Yao2019} dataset, DocRED,\footnote{\url{github.com/thunlp/DocRED}} represents both intra-sentence and cross-sentence relations. There are instances of 96 relation types from Wikidata\index{Wikidata}. The dataset has several types of annotations apart from relations: mentions, coreference links, and text fragments marked as supporting evidence for the annotated relations.

Several datasets are available for n-ary relations. There are Wiki-90k and WF-20k,\footnote{\url{github.com/aurtg/n-ary-dataset}} built from binary relation instances in Wikidata\index{Wikidata} and Freebase\index{Freebase}. \newcite{Akimoto2019} defined ternary relations by combining binary relations. Instances of these relations were mapped onto paragraphs consisting of at most three sentences from the English Wikipedia\index{Wikipedia}, processed with Stanford CoreNLP for dependency parsing\index{parsing!dependency parsing}\index{dependency!dependency parsing} and coreference resolution, and with DBpedia\index{DBpedia} Spotlight for entity detection.

\e{Few-shot learning} is a new direction of research in relation extraction\index{relation!relation extraction}/classification\index{relation!relation classification}: learning from a small number of examples. \posscite{Han2018} dataset FewRel serves this specific purpose.\footnote{\url{zhuhao.me/fewrel}} It consists of 70,000 instances, 700 instances for each of 100 relation types. The relations are derived from Wikidata\index{Wikidata} and matched with Wikipedia\index{Wikipedia} articles, and then crowd-sourced for annotation. FewRel 2.0,\footnote{\url{thunlp.github.io/fewrel.html}} developed by \newcite{Gao2019FR2}, adds a test set from the biomedical domain for exploring few-shot domain adaptation. It also provides a few-shot none-of-the-above detection setting.

A catalogue of annotated datasets for relation extraction\index{relation!relation extraction}---reference papers and links---appears in a very useful GitHub repository.\footnote{\url{github.com/davidsbatista/Annotated-Semantic-Relationships-Datasets}} %Please refer back to Section \ref{sec:data_quality} for a discussion on the quality of datasets used in relation extraction\index{relation!relation extraction}/classification\index{relation!relation classification}.

\subsection{The quality of data}
\label{sec:data_quality}

The datasets we describe in Section \ref{sec:kbs} were all built manually. There arises a natural question: how good are relation annotations in those sets? A useful answer should give the reader an idea about the difficulty of the annotation task---and possibly an upper bound on the performance of an NLP system using annotated data---as well as about the quality of annotated datasets.

Small datasets contain material for a specific experiment or task, and often arise from in-house effort which employs NLP specialists or domain experts. Inter-annotator agreement measured by some form of the kappa coefficient \cite{artstein-poesio-2008-survey} is rather regularly reported in the resulting research papers. The agreement reported for the construction process for the surveyed datasets is between 0.6 and 0.9, depending on the relation; it turns out that some relations are easier to annotate. It is a challenge to compute agreement, because, as \newcite{hendrickx2010} note, chance agreement, required in the calculation of kappa, may be difficult to estimate. Agreement on the final versions of small datasets may be moot: they often contain only adjudicated items, or items on which all annotators agree; this has been discussed, \eg by \newcite{girju09} and \citeauthor{hendrickx2010}

Larger datasets are often the result of crowd-sourced annotation. Good annotations make for a good dataset. There are methods of measuring, and keeping track of, the quality of annotations. Some form of quality control is embedded in crowd-sourcing platforms such as Amazon Mechanical Turk or Figure Eight (formerly known as CrowdFlower). They allow the task organizers to evaluate candidate workers before and during the main annotation task. After reading the annotation guidelines, workers are tested on a small set of annotated instances, and those who score low can be rejected. Annotated instances can also be sprinkled throughout the task to monitor workers' performance. Information thus gathered goes into the reliability scores which accompany an annotator's work. Annotation task organizers can also plan their own quality control, for example by duplicating instances to help quantify the consistency of each rater's annotation.

Annotations obtained on a large scale from the public at large are only part of the solution. To build a reliable dataset from individual annotations is itself a complex problem \cite{Qing2014}. Despite best effort, subsequent analysis may still uncover errors and inconsistencies. \newcite{Alt2020} analyze the quality of relation labels in the TACRED dataset, and corrected instance labels in the development and test splits. They also show that such corrections affect model evaluation significantly.

And then there are repositories of collective knowledge, very large indeed. Annotation information for those datasets does not exist as such: they were built as repositories of collective knowledge, and not specifically (and perhaps somewhat narrowly) for scientific study. NLP adopted such datasets enthusiastically, but once they became the object of scientific study, their quality had to be assessed. \newcite{Farber2018} present an in-depth analysis of Freebase\index{Freebase}, DBpedia\index{DBpedia}, Wikidata\index{Wikidata}, OpenCyc\index{Cyc} and YAGO\index{YAGO}, and give many references to additional work concerned with evaluating the quality of knowledge repositories\index{knowledge repository}. Based on a sample of instances, these resources are judged in terms of accuracy, trustworthiness, consistency and other such measures. \citeauthor{Farber2018} also propose a framework for selecting the most suitable knowledge graph\index{graph!knowledge graph (KG)} for a given setting based on the computed measures.

People who judge linguistics phenomena can rarely be in complete agreement. Even so, while a dataset may never be perfect, someone has built it in order to give the community a \e{reliable} resource. In the end, then, one must live with imperfect annotations, and factor their flaws into any honest analysis of research results.

\subsection{Distant supervision}
\label{sec:distsupnn}

Distant supervision\index{learning!distant supervision} is a popular method of acquiring additional (large amounts of) training data starting with (a small set of) annotated data from some related tasks. For relation extraction\index{relation!relation extraction} in particular, large amounts of automatically annotated data---in the form of sentences with source and relation targets marked---can be obtained using out-of-context \tripleSRT\ relation triples in knowledge repositories\index{knowledge repository}. The sources and targets in these triples are mapped onto unstructured texts. The assumption is that all or most of the newly found sentences will carry the target relation. This naturally produces noisy data. Dealing with noise---or reducing it during the data generation process---is a thorny problem. There are several traditional methods of countering it. The switch to deep learning has led to new solutions of this problem; we survey them in this section.

\subsubsection{Structured learning}
\vspace{2mm}

Evidence that a sentence contains an instance of a targeted relation\index{relation!targeted relation} can come from the sentence itself or from a larger corpus. One can filter out false positives by establishing similarity between the phrase which connects potential relation arguments in a corpus and the name of the relation in the knowledge base\index{knowledge base (KB)} \cite{Ru2018}. The evidence from the sentence and from the corpus can be further aggregated to induce latent variables helpful in predicting if a relation has an instance in the given sentence \cite{Hoffmann2011}. Such latent variables which model relational information can be induced from a low-dimensional representation of a sentence produced by a convolutional neural network\index{convolutional neural network (CNN)} \cite{Bai2019}.

An entity pair from a knowledge graph\index{graph!knowledge graph (KG)} can be connected by relations of several types. Relation learning in such cases is therefore often treated as a multi-instance multi-label learning problem \cite{Hoffmann2011,Surdeanu2012}. The labels themselves, \ie the relations, can also have semantic connections. For example, the Freebase\index{Freebase} relation \rel{/location/location/capital} connecting a capital city with its country is subsumed by \rel{/location/location/contains}. Information of this kind can be harnessed to learn the filtering of the automatically annotated sentences.

An efficient way of filtering automatically generated data is to filter sets---usually bags---of instances rather than individual instances. Automatically annotated sentences can be grouped in bags in various ways. For example, a bag may contain all sentences extracted for a given relation triple. \newcite{Zeng2015} learn such a filter; their objective function applied at the bag level incorporates the uncertainty of the instance labels. The function assigns each bag a positive label (the bag has at least one positive instance) or a negative label (the bag has no positive instances). \newcite{Su2018} build an encoder-decoder model for each bag to predict a sequence of relations---starting with the most specific one---instantiated in the bag. The encoder produces a semantic representation\index{semantic representation} of the whole bag of instances; to do that, it considers a representation of the source and target entities, and a semantic representation\index{semantic representation} of the sentences assembled using a CNN\index{convolutional neural network (CNN)}. The decoder is a neural model which learns dependencies between the semantic relations in the entire set from the representation of the bags. For each input bag representation, it produces a sequence of relation predictions, starting with the most specific relation which can be instantiated in the sentences in the bag.

\begin{wrapfigure}{r}{0.5\textwidth}
 \centering
 \includegraphics[scale=0.6]{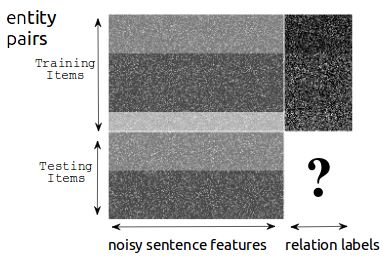}
 \caption{Entity-pair$\times$sentence matrix in distant supervision\index{learning!distant supervision} \cite{Fan2014}.}
 \label{fig:distsupmatrix}
\end{wrapfigure}

Distant supervision\index{learning!distant supervision} can be seen as the filling of entries in the label section of an entity-pair$\times$sentence co-occurrence matrix---see Figure \ref{fig:distsupmatrix}. The matrix combines gold-standard training instances and automatically labelled instances: rows represent entity pairs, columns represent (noisy) textual features from the corresponding sentences and (incomplete) relation labels.

 To fill in the incomplete relation labels, the matrix is factorized into two low-rank matrices: item$\times$feature and item$\times$label. The assumption is that the noisy features and the incomplete labels are semantically correlated \cite{Fan2014}. The resulting low-dimensional feature and label representations can help compute the relation labels for the test data.

A knowledge graph\index{graph!knowledge graph (KG)} (KG)---the usual source in distant supervision\index{learning!distant supervision}---provides much more information than just individual relation triples. \newcite{Wang2018distsup} do not use the labels associated with automatically extracted sentences. Instead, they devise a relation-learning process which relies on the fact that multiple entity pairs from a knowledge graph\index{graph!knowledge graph (KG)} communicate the same relations, and that some pairs may appear in only one kind of relation (\eg the relation between Toronto and Canada can be \rel{/location/location/contains} but not \rel{/location/location/capital}); there also is information about the type of entities connected by a given relation and an encoding of the KG\index{graph!knowledge graph (KG)} relation. From the relation triples in the KG, the system induces entity and relation representations using the TransE model. The model approximates each relation type as a translation vector in a low-dimensional space: $source + relation \approx target$ (this will be shown in Table \ref{tab:graphembs} in Section \ref{sec:linkpred}). In the extracted sentences, source and target entities are replaced with their types as supplied by the KG\index{graph!knowledge graph (KG)}. A neural network with attention\index{attention} learns sentence embeddings\index{embedding!sentence embedding} such that the embedding of a sentence is close to the target-source pair, so ultimately close to the representation of a relation. At test time, a sentence is assigned a relation label dictated by its embedding and its closest relation induced with TransE.

\newcite{Vashishth2018} also take advantage of entity type information from Freebase\index{Freebase} and relation alias information---different relation names in \tripleverb\ triples extracted from texts---to impose soft constraints on relation prediction. A graph convolution network\index{graph convolution network (GCN)} formalism is used to encode syntactic information from candidate sentences and to produce sentence embeddings\index{embedding!sentence embedding}. From these representations, the system induces representations of bags of instances, which are then combined with relation embeddings\index{embedding!relation embedding}\index{relation!relation embedding} and entity type embeddings\index{embedding!word/entity embedding}, and a softmax classifier predicts the relation.

\subsubsection{Adversarial networks}
\vspace{2mm}

Generative adversarial networks (GANs)\index{generative adversarial network (GAN)}\index{learning!adversarial learning} have had much success in dealing with the lack of training data, because they automatically generate new data to match a small set of gold-standard annotated data. In this formalism, a generator is pitted against a discriminator. The generator tries to generate data according to an underlying (unknown) distribution, and the discriminator tries to distinguish automatically generated data from the (relatively little) gold standard data with the desired distribution. The generator works best when the discriminator fails; this shows that it can generate data from the desired distribution. While this idea is not new \cite{Schmidhuber1999}, and has been implemented for traditional learning paradigms \cite{Dalvi2004,Zhou2012}, it has proven particularly fruitful in deep learning \cite{Goodfellow2014}.

The training of a generator and a discriminator has been applied in filtering instances for the distance supervision of relation classification\index{relation!relation classification}. Here, the generator need not actually generate new instances; it can just sample from the set of sentences automatically annotated for relations. The problem is to sample the true positives from the automatically generated noisy data. The discriminator is tested on a small amount of gold-standard annotated data. According to \citeauthor{Goodfellow2014}, the process ``wins'' when the discriminator cannot distinguish between its gold-standard true positives and the sentences selected from the automatically annotated data.

The sampling of true positives can be based on a computed probability that a given sentence contains an instance of the target relation \cite{Qin2018}; the sentences with the highest probability are passed on to the discriminator. Normally, a GAN's discriminator would pitch the automatically selected instances against a gold-standard annotated set. \citeauthor{Qin2018} forgo supervision. They assume a rough split of the automatically annotated data, based on the overlap with the source of distant supervision\index{learning!distant supervision}. A set $P$ of ``true positive'' data comprises sentences with both arguments of an existing relation in Freebase\index{Freebase}. In the sets $N^G$ and $N^D$ (negative data for the generator and the discriminator, respectively), the entities in the sentence do not appear in a relation instance in Freebase\index{Freebase}.

\citeauthor{Qin2018} use $P$ and $N^G$ to pretrain a generator model; the model is updated until a stopping criterion has been met, as in typical GANs. Unlike a typical GAN, the discriminator is also pretrained, and this configuration is restored at the beginning of each epoch. The generator assigns a probability score to every instance in $P$. The set of instances is split into $T$ (instances with a high probability) and $F = P \setminus T$. The parameters of the discriminator are adjusted in such a way that $T$ plays the role of negative data, $F$ of the positive data. The loss function of the discriminator computes a signal which is only used to determine the reward function for adjusting the parameters of the generator. The performance of the discriminator is evaluated on $N^D$, the set of negative examples. The intuition seems to be as follows: if the discriminator learns to assign lower probability to instances of $T$ (which it treats as negative), then it will become worse at distinguishing positive and negative instances, so it will perform poorly at scoring the negative data $N^D$.

\begin{wrapfigure}{r}{0.5\textwidth}
 \centering
 \vspace{-3mm}
 \includegraphics[scale=0.4]{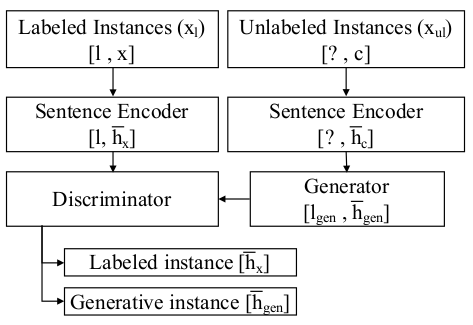}
 \caption{\posscite{Li2019} GAN architecture: the same encoder for gold-standard and automatically annotated sentences.}
 \label{fig:GAN}
\end{wrapfigure}

Information in a sentence often ends up compressed into a single probability value. Such compression obscures the various facets of the sentence. There is an alternative: construct a vector representation of the sentence, perhaps using a convolutional neural network\index{convolutional neural network (CNN)}; next, use it to build and train the generator and the discriminator \cite{Li2019}. This procedure, illustrated in Figure \ref{fig:GAN}, can be combined with additional methods of filtering the automatically annotated data. \citeauthor{Li2019} produce a cleaner training set from entity descriptions collected from Wikipedia\index{Wikipedia}. As positives instances, they take the sentences with entity mentions which appear not only in a relation in the source of distant supervision\index{learning!distant supervision}, but in each other's description. If they do not, the sentence is considered a negative instance.

\subsubsection{Neural networks with attention\index{attention}}
\vspace{2mm}

Automatically generated true positive and false positive sentences may share features which can be exploited to filter the distantly supervised\index{learning!distant supervision} dataset. To find such features, the meaning of the sentences should be represented in a systematic manner, by a mechanism which reveals shared patterns\index{pattern}. 

Convolutional neural networks\index{convolutional neural network (CNN)} are good at finding patterns. The induced sentence representations can be used directly with a sentence-level attention\index{attention} model which reduces the weights of noisy (false positive) sentences \cite{Lin2016}. They can also lead to aggregate representations of groups of sentences, in particular bags of sentences extracted for each relation $r$ and $(source, target)$ pair.

\newcite{Ye2019} apply a bag-level attention\index{attention} mechanism to relation-aware representations built for each bag as (attention\index{attention}-)weighted sums of sentence representations matched against each relation. Sentence representations are built by CNNs\index{convolutional neural network (CNN)} over word embeddings\index{embedding!word/entity embedding}, taking into account positional information about the source and the target. This weighted sum of sentence representations is matched against every possible relation, not just the target relation: the same entity pair can have different relations in different contexts (\eg \e{\dul{Barack Obama} was born in the \dul{United States}} and \e{\dul{Barack Obama} was the 44th President of the \dul{United States}}). Bags which share a relation label are assembled into a bag group. An attention\index{attention} mechanism helps weight sentences for the construction of the bag representation. The mechanism is expected to give smaller weight---pay less attention---to noisy sentences. A similar attention\index{attention} mechanism should give lower weight to noisy bags in a bag group. The attention\index{attention} model is trained to weight more highly sentences more likely to express the desired relations.

\newcite{Beltagy2019} combine adversarial training\index{learning!adversarial learning} with attention\index{attention}; that improves the automatic selection of positive instances from automatically selected---therefore noisy---sentences which contain the target relations. \citeauthor{Beltagy2019}'s work improves the model's ability to assign lower weights to noisy sentences, those which do not contain the target relation.

Another trouble with distant supervision\index{learning!distant supervision} is that sentences contain more information than just the target relation, and such additional phrases may conceal the target relation. Filtering out some of the noise can make it easier for an attention\index{attention} model to find, and properly weight, relevant features. \newcite{Liu2018} implement such a filter. In word-level distant supervision\index{learning!distant supervision} for relation extraction\index{relation!relation extraction}, where filtering is based on syntactic information, a robust entity-wise attention\index{attention} model will give more weight to semantic features of relational words in a sentence.

Attention\index{attention} models are usually implemented as weight vectors. This one-dimensional view may insufficiently account for complex interactions in textual contexts. \newcite{Du2018} propose a multi-level multi-dimensional attention\index{attention} model in a multi-instance learning framework. A 2D attention\index{attention} matrix identifies aspects of the interaction of two entities in a sentential context, while another 2D attention\index{attention} matrix picks up relevant sentence-level features.

The attention\index{attention} mechanism relies on the neural model for the encoding of relevant and noteworthy features. With respect to relation extraction\index{relation!relation extraction}, some such features should capture semantic aspects of the entities involved. The sentential context may not contain enough information for this, so one could give the model additional information in the form of entity descriptions. \newcite{Ji2017} extract entity descriptions from Freebase\index{Freebase} and Wikipedia\index{Wikipedia} pages, and give those to a model which includes sentence-level attention\index{attention}.

\subsubsection{Reinforcement learning}
\vspace{2mm}

The decision which instance is useful\index{learning!reinforcement learning}---is a true positive---can be treated as a game. In a game, generally speaking, an agent starts in an initial state, chooses a series of actions which lead to a final state, and is rewarded or penalized depending on whether the final state is good or bad. This reward/penalty controls the adjustment of the agent's model, which dictates what actions to take in a given state.

In relation extraction\index{relation!relation extraction}, the purpose of the game is to find true positive sentences among those automatically extracted and annotated. At each step, the agent chooses an instance from this automatically generated set. When it reaches the final state, the set of instances gathered is considered to be a training set of properly labelled instances. A model is built from this training set, and evaluated on a small set of gold-standard data. A good training set yields a good model for learning to predict relations, and a noisy training set leads to poor performance on the task. The result of this evaluation determines the agent's reward or penalty; based on that, the agent adjusts its method of choosing instances and assigning each of them a positive or negative label.

Formally, \e{state} $s_t$ corresponds to the training data selected until time $t$, the target relation $r_i$, and the relation arguments---$source$ and $target$---in the currently considered instance. Three \e{actions} are possible: accept it as true positive and include it in the training data; include it as a negative instance; or reject it. The \e{reward} is computed after the instances in the automatically generated dataset (or in a bag corresponding to a given $source\text{-}target$ pair) have been processed, and the final relation classification\index{relation!relation classification} model built and evaluated. 

\begin{wrapfigure}{r}{0.5\textwidth}
 \centering
 \includegraphics[scale=0.3]{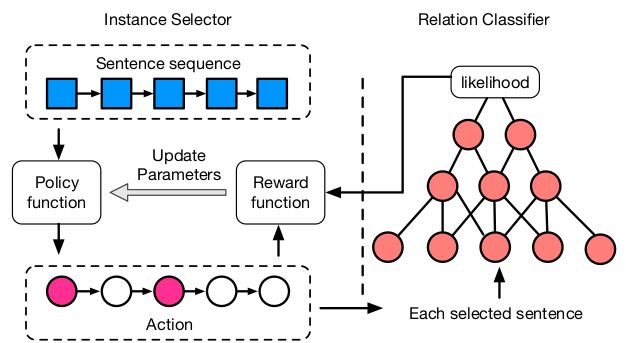}
 \caption{Example of a reinforcement-learning architecture in which a relation classifier\index{relation!relation classification} selects instances \cite{Feng2018}.}
 \label{fig:RI}
\end{wrapfigure}

\ \newcite{Feng2018} and \newcite{Qin2018RL} apply this form of reinforcement learning\index{learning!reinforcement learning} in the selection of training data for relation learning. When the agent decides on the action, it relies on a relation classification\index{relation!relation classification} model, implemented as a convolutional neural network\index{convolutional neural network (CNN)}. The reward/penalty feedback guides the adjustment of the parameters of this CNN\index{convolutional neural network (CNN)}. Figure \ref{fig:RI} illustrates.

The process can be improved if the focus is not only on false positives but on false negatives. (A false positive is an automatically generated sentence which does not actually have one of the target relations, despite containing both arguments of a relation. A false negative is a sentence discarded because it did not contain an exact expression of the arguments of a relation.) \newcite{Yang2019} focus on identifying both false positives and false negatives. They treat this as a reinforcement problem, in which an agent should decide if an instance is mislabelled. The task is split between two agents: one of them detects false positives, the other false negatives.

The adjustment of action selection as a result of feedback is a critical step in reinforcement learning\index{learning!reinforcement learning}. Earlier work relied on a relation classifier\index{relation!relation classification} to take an action, and the reward/penalty indicated how to adjust the parameters of the model. The action can also be the result of more complex processing. \newcite{Liu2019} developed a reinforcement learning\index{learning!reinforcement learning} process in which a GAN-like method performs a \e{policy} improvement step. A policy is a probability distribution which maps states to actions. The improvement of a policy was modelled as a form of imitation learning.\footnote{Imitation learning aims to mimic human behavior in a given task. An agent is trained to perform the task from demonstrations by learning a mapping between observations and actions.} It takes the current policy as the prior knowledge, and generates improved policies. The reward is implicit in the policy improvement operator. In the policy evaluation step, the current policy network is rated by a measure of the difference between the probability distribution under the current policy and under the improved policy.

%% %% section %% %%
\section{Learning semantic relations}
\label{sec:relationsDL}

There are many ways of representing the semantics of words/entities, relational clues from the context, and even the relations themselves. Various architectures can combine this information, and in fact it is often derived jointly. This section presents several ways in which such representations can be integrated in a deep-learning method. The methods are grouped by the sources and information support for the relation learned: learning relations in knowledge graphs\index{graph!knowledge graph (KG)} (Section \ref{sec:graphDL}), learning relations from texts (Section \ref{sec:textDL}), learning relations from texts and knowledge graphs\index{graph!knowledge graph (KG)} (Section \ref{sec:textgraphDL}), n-ary and cross-sentence relations (Section \ref{sec:naryDL}), unsupervised\index{learning!unsupervised learning} relation extraction\index{relation!relation extraction} (Section \ref{sec:unsupDL}), and finally lifelong learning (Section \ref{sec:lifelongDL}).

\subsection{Learning relations in knowledge graphs}
\label{sec:graphDL}

A knowledge graph\index{graph!knowledge graph (KG)} $KG = (\mathv,\mathe,\mathr)$ contains knowledge in the form of relation triples $(s,r,t)\in\mathe$, where the vertices $s,t \in \mathv$ are entities and $r \in \mathr$ is a relation type. Knowledge graphs\index{graph!knowledge graph (KG)} are not complete. Additional links (facts) can be inferred because similar nodes participate in similar relations, \eg every country has a capital city. This is, in effect, \e{link prediction}\index{link prediction}. To do this, a KG may be encoded by a learning model which approximates its connectivity information; or one can apply a graph neural network\index{graph neural network (GNN)} to encode its structure via node neighbourhoods.

\subsubsection{Encoding graph connectivity information}
\label{sec:linkpred}
\vspace{2mm}

Graph embedding\index{embedding!graph embedding} methods rely on the idea that the graph's connectivity structure informs the representations of entities and relations. The representation of an entity takes into account the relations it is part of; the representation of a relation takes into account the entities it connects. A few essential design decisions must be made: the type of structure to represent entities and relations (\eg vectors or matrices); the scoring function\index{scoring function} to calculate a score for a pair of entities and a relation type based on these representations (the score should be 1 if the edge in question exists); and the loss function to compare the automatic predictions with the gold standard. \newcite{Nickel2016b} present a survey of statistical relational learning methods for knowledge graphs\index{graph!knowledge graph (KG)}, including graph embeddings\index{embedding!graph embedding}, path-based algorithms and Markov logic networks. \newcite{Wang2017} and \newcite{Ji2020} focus on knowledge graph\index{graph!knowledge graph (KG)} embedding\index{embedding!graph embedding} methods, and present a comprehensive overview. This section summarizes some of those methods; the surveys offer the reader a deeper look.

Knowledge graphs\index{graph!knowledge graph (KG)} contain only positive instances, \ie relation instances known to exist. The link prediction\index{link prediction} task precludes the closed-world assumption, otherwise, every missing link would be a legitimate \rel{not\_related} relation. To produce non-trivial models, ``negative'' edges are needed. There is variety of methods which sample a number of missing edges for each positive instance \cite{Kotnis2018}. A scoring function\index{scoring function} applied to such apparently negative instances should return a score close to 0. Alternatively, since these instances are only presumed to be negative, the scoring and the loss functions can implement \e{contrastive estimation} \cite{Gutmann2012}: the score for a positive instance should be higher that the score for all the negative instances sampled for it.

Graph embedding\index{embedding!graph embedding} methods learn representations $\emb{v}_x$ for entity $x$ and \emb{r} for relation $r$. The fact that relation $r$ holds between the source and target nodes $s$ and $t$ is modelled by a scoring function\index{scoring function} $f$ (discussed briefly in Section \ref{sec:wefromgraphs}). The entity embeddings\index{embedding!word/entity embedding} $\emb{v}_x$ are usually $d$-dimensional vectors, where $d$ is a parameter. The representation of the relation has taken various forms, \eg a $d$-dimensional vector (Trans* \cite{Bordes2013,Wang2014,Lin2015}), a diagonal matrix (DistMult \cite{Yang2015}), or a $d\times d$ matrix (\rescal\ and its variations \cite{Nickel2011}).

The entity and the relation embeddings\index{embedding!relation embedding}\index{relation!relation embedding} can be considered to belong to different embedding spaces\index{embedding!embedding space}, and projections can map entity embeddings\index{embedding!word/entity embedding} onto the relation space \cite{Lin2015,Ji2015,Ji2016}. Entities and relations are most commonly modelled as deterministic points or vectors in continuous vector spaces. In contrast, \newcite{He2015} and \newcite{Xiao2016} propose models which represent both entities and relations as vectors drawn from Gaussian distributions. Such representations allow variations in the meaning of a semantic relation for different (\e{source, target}) pairs, and for sources and targets in different contexts.

\begin{table}[ht]
 \caption{A small sample of graph embedding\index{embedding!graph embedding} methods. There is a comprehensive overview in \cite{Wang2017} and \cite{Ji2020}. {\small References for the methods: TransE \cite{Bordes2013}, DistMult \cite{Yang2015}, \rescal\ \cite{Nickel2011}, ComplEx \cite{Trouillon2017},
 TransG \cite{Xiao2016}, ConvE \cite{Dettmers2018}, MuRP \cite{Balazevic2019}}.}
 \label{tab:graphembs}
 \plaincenterline{\mytable
{\small
 \begin{tabular}{cccc} 
 \hline 
 \cb {\bf Method} 
 & \cb {\bf Entity emb.} 
 & \cb {\bf Relation emb.} 
 & \cb {\bf Scoring function\index{scoring function}} \\
 \hline
 \noalign{\vskip 0.1in}
 %%%%%%
 TransE 
 & $\embv_s,\embv_t \in \realr^d$ 
 & $\embr \in \realr^d$ 
 & $\norm{\embv_s + \embr - \embv_t}$ \\
 \noalign{\vskip 0.1in}
 %%%%%%
 \cy DistMult 
 & \cy $\embv_s,\embv_t \in \realr^d$ 
 & \cy $\embr \in \realr^d$ 
 & \cy $\embv_s^\transp diag(\embr) \embv_t$ \\
 \noalign{\vskip 0.1in}
 %%%%%%
 \rescal 
 & $\embv_s,\embv_t \in \realr^d$ 
 & $\emb{M_r} \in \realr^{d\times d}$ 
 & $\embv_s^\transp \emb{M_r} \embv_t$ \\
 \noalign{\vskip 0.1in}
 %%%%%%
 \cy ComplEx 
 & \cy $\embv_s,\embv_t \in \realc^d$ 
 & \cy $\embr \in \realc^d$ 
 & \cy $Re(\embv_s^\transp diag(\embr) \bar{\embv_t})$ \\ \hline
 \noalign{\vskip 0.1in}
 %%%%%%
 TransG
 &
 \parbox{1.0in}{\centering $\embv_s \sim \gauss(\mu_s,\sigma_s^2I)$ \\ $\embv_t \sim \gauss(\mu_t,\sigma_t^2I)$ \\ $\mu_s, \mu_t \in \realr^d$ \\ $\Sigma_s,\Sigma_t \in \realr^{d\times d}$}
 &
 \parbox{1.65in}{\centering $\mu_r^i \sim \gauss(\mu_t - \mu_s, (\sigma_s^2 + \sigma_t^2)I)$ \\ $\embr = \sum_i \pi_r^i \mu_r^i \in \realr^d $ \\ {\scriptsize ($\pi_r^i$ are weights)}}
 &
 \parbox{1.65in}{$\sum_i \pi_r^i exp\left(- \frac{\norm{\mu_s + \mu_r^i - \mu_t}_2^2}{\sigma_s^2 + \sigma_t^2}\right)$} \\ \hline
 \noalign{\vskip 0.1in}
 %%%%%%
 \cy ConvE
 &
 \cy \parbox{1.0in}{\centering $\embv_s \in \realr^{d_h\times d_w}$ \\
 $(d_h d_w = d)$ \\ $\embv_t \in \realr^d $ }
 &
 \cy \parbox{1.65in}{\centering $\embr \in \realr^{d_h\times d_w}$ }
 &
 \cy \parbox{1.65in}{$f(vec(f([\embv_s ; \embr ] \ast \omega))W)\embv_t$ \\
 {\scriptsize ($\ast$ is the convolution operator, \\
 $\omega$ is a CNN filter, \\
 $W$ is a weight matrix)}
 } \\ \hline
 \noalign{\vskip 0.1in}
 %%%%%%
 MuRP 
 &
 \parbox{1.0in}{\centering $\emb{h}_s,\emb{h}_t \in \bball_c^d$ \\
 $\bball_c^d = \{x \in \realr^d : c\norm{x}^2 < 1 \}$ }
 &
 \parbox{1.65in}{\centering $\emb{r} \in \bball_c^d$ \\
 $\emb{R} \in \realr^{d\times d}$ }
 &
 $ - d_{\bball}(\emb{h}_s^{(r)},\emb{h}_t^{(r)})^2 + b_s + b_t$ 
 \\ \hline
\end{tabular}
} % small
\emytable}
\end{table}

Table \ref{tab:graphembs} shows examples of graph embedding\index{embedding!graph embedding} models. The column \e{Entity emb.} contains the implementation choice for the entity embeddings\index{embedding!word/entity embedding}: a real- or complex-valued $d$-dimensional vector, or a $d$-dimensional vector drawn from a normal distribution.\footnote{$\gauss(\mu,\sigma^2I)$ represents the normal distribution with mean $\mu$ and covariance matrix $\sigma^2I$.} The column \e{Relation emb.} lists the chosen representation structure for the relation: a real- or complex-valued $d$-dimensional vector, a real-valued $d\times d$-matrix, or a $d$-dimensional real-valued vector drawn from a normal distribution. The \e{Scoring function}\index{scoring function} column presents the calculation of the score for a relation triple, given the representation choices.

Methods such as \rescal\ \cite{Nickel2011} and Neural Tensor\index{tensor} Networks \cite{Socher2013} learn millions of parameters. That makes them more flexible---they can model a variety of relations well---but there are costs: increased computational complexity and a chance of overfitting. Methods such as TransE \cite{Bordes2013} and DistMult \cite{Yang2015} learn simpler models, with far fewer parameters, and are easier to train, but they cannot model certain types of relations, such as many-to-many relations (TransE) and asymmetric relations (DistMult). \newcite{Nickel2016}'s holographic embeddings\index{embedding!graph embedding} (HolE) achieve the modelling power of \rescal\ with fewer parameters by compressing the tensor\index{tensor} product. Complex-valued embeddings\index{embedding!graph embedding} (ComplEx) \cite{Trouillon2017} extend DistMult to model antisymmetric relations.

\newcite{Nickel2017} first proposed embedding (part of) \wn's\index{WordNet} \isa\index{relation!is-a relation} hierarchy in a Poincar\'{e} space; their hyperbolic embeddings\index{embedding!hyperbolic space embedding} (of very low dimension: $d$ = 5) predicted unseen \isa\ instances. \posscite{Balazevic2019} model embeds and predicts links in a multi-relational knowledge graph\index{graph!knowledge graph (KG)}. Their MuRP model, illustrated in Table \ref{tab:graphembs}, represents entities as points in a Poincar\'{e} ball. A scoring function\index{scoring function} determines if two entities are in a relation $r$. The function relies on relation-specific parameters $\emb{r}$ (a hyperbolic translation vector) and $\emb{R}$ (a diagonal relation matrix). The parameters transform the source and target hyperbolic embeddings\index{embedding!hyperbolic space embedding} $\emb{h}_s$ and $\emb{h}_t$ into $\emb{h}_s^{(r)},\emb{h}_t^{(r)}$. Two biases, $b_s$ and $b_t$ (which are among the model's parameters) define a ``sphere of influence'' around each of the transformed vectors. If $r$ connects the source and target entities, then their spheres of influence should overlap.

A graph-structure encoding approximates, in effect, the adjacency matrix\index{adjacency matrix} of a graph. The matrix captures the view of a graph as a collection of triples. A graph can also be represented as a collection of paths. Paths in graphs can result from graph traversal (breadth-first, depth-first, and so on) or random walks. 

Paths can assist relation learning in various ways. When regarded as a sequences of nodes and relations, a path can serve as a ``sentence'' for the purpose of deriving node and relation representations. Paths are the input, in lieu of a regular corpus, to word2vec \cite{mikolov2013efficient}, and representations of the nodes and relations are produced just as one would do it for words in a sentence \cite{Perozzi2014}.

A path can be treated as a description of the source and target nodes (it contains information about their neighbourhoods) or the relation between them (it shows alternative chains of links from the source to the target). Paths, then, can contribute directly as features to the prediction of new links in a knowledge graph\index{graph!knowledge graph (KG)}. \newcite{Lao2011} show how to obtain and apply bounded-length path types, or meta-paths (sequences of relations): they generalize alternative paths found between the source and the target in the graph connected by the same relation $r$. The meta-paths work as features in predicting if relation $r$ holds between node pairs previously not connected by $r$.

\newcite{Gardner2015} use paths to describe the source and target nodes, and the relation between them. They extract features from the local subgraphs around each node in a potential pair. The local information around node $n$ is the set of $($\e{path type}, \e{end node}$)$ pairs collected by random walks which originate in $n$. The representation for a ($source$, $target$) pair combines the two nodes' subgraphs by merging the paths based on shared end nodes on those paths. This representation is used to learn a corresponding relation for the entity pair.

\newcite{Guu2015} show that most latent factor models, notably matrix factorization\index{matrix factorization} models, can be modified to learn from paths rather than from individual triples. Recurrent neural networks\index{recurrent neural network (RNN)} which learn path representations have also been used for link prediction\index{link prediction} \cite{Neelakantan2015,Das2016}.

Relations can also share information. For example, the relation \rel{currency\_of\_film\_budget} can be viewed as a composition of the relations \rel{currency\_of\_country} and \rel{country\_of\_film}. This kind of information may promote better relation representations. \newcite{Takahashi2018} use an autoencoder which further processes the relation matrices obtained by matrix factorization\index{matrix factorization} with the \rescal\ model. The autoencoder compresses the relation matrix into a smaller vector representation, from which the matrix is regenerated. This encoding-decoding process encourages the induction of relation matrices which incorporate similarities and dependencies between the relations.

The previously described graph embeddings\index{embedding!graph embedding} took into account the structure of the graph, and encoded entities and relations in various types of structures. The entity and relation representations determine the scoring function\index{scoring function} used to approximate the graph structure and then to predict new edges.

Information in a knowledge graph\index{graph!knowledge graph (KG)} can be encoded in other ways, for example when relation triples are taken into account as separate instances. The focus in such a case would be on modelling the interaction between arguments and relations to boost latent patterns, such as shared or interacting dimensions. \newcite{Dettmers2018} use a multi-layer CNN\index{convolutional neural network (CNN)}, whose input is a 2D encoding of the source entity and relation in a \tripleSRT\ triple. The filters applied to this source-relation ``image'' are common to all instances in the training data, and so to all relation types. The application of the filters over the 2D representation produces feature maps; a fully connected layer projects the maps onto a hidden layer which represents an entity embedding\index{embedding!word/entity embedding}. The predicted embedding vector is multiplied with the entity matrix and then transformed by a sigmoid function; that, in effect, produces a similarity score between the predicted embedding and the embeddings of the entities in the graph. \citeauthor{Dettmers2018}'s method makes it possible to perform a 1-to-N mapping, simultaneously testing all possible targets of a source-relation combination.

\newcite{Jiang2019} take \citeauthor{Dettmers2018}'s method further. They start from the premise that concatenating the 2D representations of the subject and the relation does not allow for enough interaction between the dimensions of the subject and relation. \citeauthor{Jiang2019}'s system takes as input only a 2D representation of the source entity, and the representation of the relation is transformed into a set of filters. This enables a more diverse and comprehensive interaction between the representation of the subject and the relation. In contrast with \citeauthor{Dettmers2018}'s work, each relation type has its own filters. 

\subsubsection{Graph Neural Networks}
\label{sec:gnnskg}
\vspace{2mm}

The graph encoding methods which we discussed in connection with link prediction\index{link prediction} do not take full advantage of the graph structure. For example, a node's neighbourhood should provide useful information. Graph neural networks (GNNs)\index{graph neural network (GNN)} were designed to acquire such information down to any depth. GNNs\index{graph neural network (GNN)}, proposed first by \newcite{Scarselli2009}, aggregate this information into a fixed-sized representation. The aggregation function---message passing---must be invariant in the neighbourhood shape or size. \newcite{Zhou2018} present a comprehensive view of GNNs\index{graph neural network (GNN)}; it is summarized here from the point of view of their connection to semantic relation learning.

GNNs\index{graph neural network (GNN)} were inspired by convolutional neural networks\index{convolutional neural network (CNN)}, which can find patterns\index{pattern} at different levels and then compose them into expressive representations. There are three key aspects of CNNs\index{convolutional neural network (CNN)} which allow them to produce such representation: local information (in graphs, it is node neighbourhood); shared weights (this reduces the computing cost); and multi-layer structures which deal with hierarchical patterns\index{pattern} and so capture features of various sizes (this maps naturally into the hierarchical structure of graphs). It is an important characteristic of a GNN\index{graph neural network (GNN)} that its output is invariant in the input order of nodes. The relation information, which represents the dependency\index{dependency!dependency relation} between two nodes, can be explicitly integrated into the model, including the relation's potential attributes. 

A node is defined by its features, the related nodes, and the type of relations which connect it with its neighbours. Learning a GNN\index{graph neural network (GNN)} implies learning a hidden state $\emb{h}_v \in \realr^d$ which encodes the neighbourhood information for node $v$. This vector can be used to produce an output $\emb{o}_v$ which corresponds, for example, to $v$'s label. The basis for inducing such representations is a \e{local transition function} $f$ which combines the features of the node ($\emb{x}_v$), the features of its edges ($\emb{x}_{co[v]}$), the states of the nodes in its neighbourhood ($\emb{h}_{ne[v]}$), and the features of the nodes in its neighbourhood ($\emb{x}_{ne[v]}$). Formally, the hidden state is defined as
\begin{align}
\emb{h}_v = f(\emb{x}_v,\emb{x}_{co[v]},\emb{h}_{ne[v]},\emb{x}_{ne[v]})
\end{align}
The output depends on $v$'s hidden state and feature vector\index{feature!feature vector}. It is defined as
\begin{align}
\emb{o}_v = g(\emb{h}_v,\emb{x}_v)
\end{align}
where g is a local output function. To learn $g$'s and $h$'s internal parameters, GNNs\index{graph neural network (GNN)} need a loss function. It compares the predicted output $\emb{o}_v$ for a node with the gold-standard $\emb{t}_v$ from a given training set $\mathcal{V}$:
\begin{align}
loss = \sum_{v \in \mathcal{V}}(\emb{t}_v - \emb{o}_v)
\end{align}

Not only are knowledge graphs\index{graph!knowledge graph (KG)} incomplete but they are incomplete in an imbalanced way. The node degrees and relation frequency plots for Freebase\index{Freebase} and NELL\index{Never-Ending Language Learner (NELL)} in Figure \ref{fig:kgstats} (in Section \ref{sec:wefromgraphs}) illustrate this difficulty. Because of the skewed structure, using graph neural networks\index{graph neural network (GNN)} to encode large-scale knowledge graphs\index{graph!knowledge graph (KG)} can give low-quality encodings of nodes and relations. One way of dealing with this skewness is to limit the size of the considered neighbourhood by sampling. \newcite{Niepert2016} shows how to map discriminative Gaifman models (a new family of relational machine learning models) onto KGs\index{graph!knowledge graph (KG)} by learning representations from local bounded-sized neighbourhoods. The model is built bottom up from these neighbourhoods, allowing for the efficient transfer of learned representations between connected objects.

A graph convolution network (GCN)\index{graph convolution network (GCN)} aggregates the signal for each node in the network: it sums over the incoming signals from the node's predecessor nodes. The signal can be enriched with information about the type of relations between connected nodes, making them Relational Graph Convolution Networks\index{graph convolution network (GCN)} (R-GCN). \posscite{Schlichtkrull2018} transformation of the incoming signal from a connected node is based on the connecting relation. The transformation is encoded as matrix multiplication, where each relation is represented by its own matrix. The resulting representations can be used in a link prediction\index{link prediction} formalism, as discussed in the preceding subsection.

To assist in the task of relation classification\index{relation!relation classification}/extraction\index{relation!relation extraction}, the GCN\index{graph convolution network (GCN)} formalism can be applied not only to knowledge graphs\index{graph!knowledge graph (KG)} but to graphs which identify connections between relation types. Freebase\index{Freebase} relations, for example, have specific names (such as \rel{/people/person/ethnicity} or \rel{/people/person/nationality}) which help organize the relations themselves into a graph structure. The adoption of such a relation graph as additional information in the KG encoding process encourages similar relations to have similar representations. \newcite{Zhang2019longtail} initialize the representation of the leaf relations with representations induced by matrix factorization\index{matrix factorization}, and the representations of internal nodes with averages of the representations of their children. \citeauthor{Zhang2019longtail} then use GCN to update these representations, so that similar relations have similar representations. The purpose of this process is to bootstrap additional information from the KG\index{graph!knowledge graph (KG)} to induce more informative representations for low-frequency relations, and ultimately help link prediction\index{link prediction}.

The GNN\index{graph neural network (GNN)} formalism is also particularly adept at including various types of information which express relevant features of the nodes and the edges. \citeauthor{Schlichtkrull2018}'s R-GCN\index{graph convolution network (GCN)} has shown how to integrate information about relation types in the model. \newcite{Garcia-Duran:2017} focus on node information, and include a variety of attributes, including numeric and multi-media features.

Hyperbolic spaces have been shown to capture structural properties of graphs better than their Euclidean counterparts; see, \eg \cite{Nickel2018}. That is why it is natural to consider an extension of graph neural networks\index{graph neural network (GNN)}, which exploit structural properties of graphs, to hyperbolic spaces \cite{Liu2019hyperbolic,Chami2019}. This requires tackling a few problems: map the input Euclidean node features to a hyperbolic space, perform set aggregation in hyperbolic space, and choose the hyperbolic spaces with the right curvature. \newcite{Chami2019}, building upon the GCN\index{graph convolution network (GCN)} architecture, propose hyperbolic graph convolution networks (HGCN)\index{hyperbolic graph convolution network (HGCN)}. This machinery combines the expressiveness of GCNs with hyperbolic geometry solutions of the issues of input representation and set aggregation for the problem of message-passing in GNNs. \citeauthor{Chami2019} show that embeddings\index{embedding!hyperbolic space embedding} which their HGCN learns preserve hierarchical structure. That leads to improved performance, when compared to Euclidean counterparts, on link prediction\index{link prediction} on several sets of medical data.

\subsection{Learning relations from texts}
\label{sec:textDL}

Relation learning from texts takes two forms. One focuses on relation classification\index{relation!relation classification}; it assumes that the relation arguments are given, and aims to predict the relation between them. The other is the joint learning of arguments and relations from unmarked texts.

\subsubsection{Relation classification}
\vspace{2mm}

A successful model for relation classification\index{relation!relation classification} relies on detecting shared patterns\index{pattern} across a number of instances. Convolutional neural networks (CNNs)\index{convolutional neural network (CNN)} do this particularly well. They were initially applied in image processing, and performed very well in noticing patterns distributed over various regions of an image \cite{LeCun1995}. The idea behind CNNs for relation classification\index{relation!relation classification} is to find common patterns\index{pattern} in the text which surrounds or connects instances of the same relation. The context\index{context representation}, which has varying length and complexity, can be input to the CNN\index{convolutional neural network (CNN)} in diverse ways. Some systems \cite{Liu2015,Xu2016,Can2019} rely on producing a fixed-length vector using one of the compositional\index{compositionality} methods (see Section \ref{sec:compositionality}). Another possibility is to have a fixed-size window centered on the relation arguments, or to slide it over the context and pool the representations to produce ``summaries'' of the context based on various input masks \cite{Zeng2014,Nguyen2015}. 

Once a fixed-sized vector for an input sentence has been acquired, this representation can help calculate a score for the sentence with respect to a semantic relation. The calculation is based on a vector representation for each relation type, which is also a learned parameter of the model \cite{Santos2015}.

\begin{figure}[ht]
 \centering
 \includegraphics[scale=0.36]{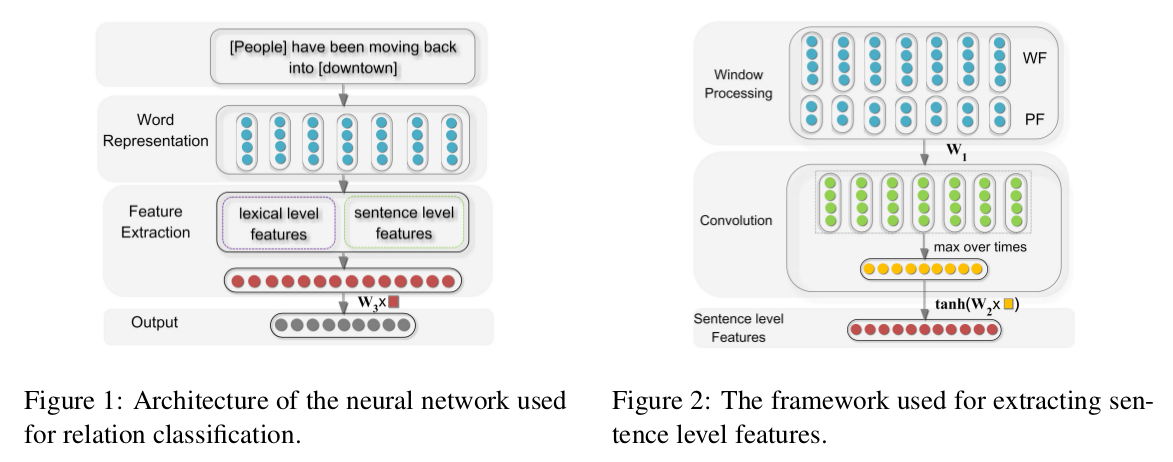}
 \caption{Learning relations using a CNN\index{convolutional neural network (CNN)} \cite{Zeng2014}.}
 \label{fig:Zeng2014CNN}
\end{figure}

The representation of a sentence can have multiple segments, to find separately information relevant to relation learning. For example, \newcite{Zeng2014} produce a representation with two sections, one for the target words, another for sentence-level features. These global features are induced by a convolutional neural network\index{convolutional neural network (CNN)} on the words of the input sentence. A word has a two-pronged representation: its embeddings\index{embedding!word/entity embedding}, and the position features which quantify its distance to the relation arguments. This is illustrated in Figure \ref{fig:Zeng2014CNN}. 

The arguments and their connecting patterns\index{pattern} can be processed separately by a CNN\index{convolutional neural network (CNN)}. The arguments could be modelled by CNNs\index{convolutional neural network (CNN)} on representations of windows of several sizes centered on those arguments. The connecting phrase can be processed similarly, by applying a CNN\index{convolutional neural network (CNN)} to the sentence fragment between the relation arguments. This leads to fixed-sized vectors representing the arguments and the relation, to be used as input to a relation classification\index{relation!relation classification} step \cite{Zheng2016}.

Dependency paths\index{dependency!dependency path} have been shown to be a useful relation indicator. Because of their varying length and structure, they require particular encoding methods. There exist compositional\index{compositionality} and graph methods for encoding such features to produce fixed-sized vectors which can serve as input to other neural networks for relation classification\index{relation!relation classification}.

Consider sentences which contain a pair of entities and are instances of the same semantic relation. The expressions of the relation in the different sentences (\eg phrases which connect the two entities) are considered mutual paraphrases\index{paraphrase}. In \posscite{Rossiello2019} work, this assumption supports the fact that if two pairs of entities represent instances of the same relation, then they are analogous. \citeauthor{Rossiello2019} compare pairs of entities
using hierarchical Siamese networks. An entity pair is represented by all the sentential contexts found for it in a corpus. The Siamese network architecture is trained to minimize the difference between entity pairs with the same relation, even when they appear in (slightly) different contexts in the corpus. In that way, it learns the different paraphrases\index{paraphrase} of the same relation.

\newcite{Shwartz2018} apply deep learning to the prediction of paraphrases\index{paraphrase} which explain noun-compound relations. They reformulate the paraphrase prediction task as three related subtasks: predict the head, the modifier, or the connecting pattern\index{pattern} (found in a corpus). This causes a tuning of pre-computed word embeddings\index{embedding!word/entity embedding} towards a state where modifier-head combinations which share similar patterns\index{pattern} are closer in the embedding space\index{embedding!embedding space}, and so are patterns\index{pattern} shared by similar modifier-head combinations.

In addition to the model architecture itself, what is essential is a good representation of the relation arguments and their contexts\index{context representation}. When a Deep Bidirectional Transformer, or BERT\index{Bidirectional Encoder Representations from Transformers (BERT)}, learns to predict targeted words in a sentence \cite{Devlin2018,devlin-etal-2019-bert}, it builds a deep representation for the entire context\index{context representation} by fusing the context on both sides of those words. Such a representation can help predict semantic relations; \newcite{Shi2019} study this hypothesis. BERT works with a masked language model: given a sentence, it masks certain words and then learns to predict them. For relation classification\index{relation!relation classification}, the word/entity arguments are masked by their grammatical role and their entity type. \citeauthor{Shi2019} show that contextualized embeddings\index{embedding!contextualized embedding} obtained in this manner predict very well the relation type between a given pair of words/entities in a sentence.

All these methods get and process only one path between a targeted pair of entities. In \posscite{Christopoulou2018} system, the connection between a pair of entities is described by all possible paths between them (of length at most $L$) in a complete graph which connects all entity mentions in a sentence. \citeauthor{Christopoulou2018} assume that entity mentions and their types are given. The directed edge connecting a pair of entities in this graph is initialized by a model which combines the representation of the entities and the context\index{context representation} around them. An iterative algorithm then builds a representation of the connection between the entities by aggregating the graph walks between them of length at most $L$. This representation is used to predict the relation type.

The \isa\index{relation!is-a relation} relation is a frequent target of relation extraction\index{relation!relation extraction}. Distributional semantic models give good results, so it is natural to ask what the improved word embeddings\index{embedding!word/entity embedding} and deep learning can bring to this task. The \isa\index{relation!is-a relation} relation can be detected from the meaning of the word themselves, or from their connective patterns\index{pattern}.

For words projected into an embedding space\index{embedding!embedding space}, the \isa\index{relation!is-a relation} relation could ideally be a linear projection from the hyponym\index{relation!hyponymy} to the hypernym\index{relation!hypernymy}, or at least there can be several such projections, depending on the characteristics of the word pairs. Consider, for example, \e{(cat, animal)} vs. \e{(table, furniture)} vs. \e{(Germany, country)}. \newcite{Fu2014} propose word embeddings\index{embedding!word/entity embedding} for the discovery of clusters\index{cluster, clustering} in the set of arguments of the \isa\index{relation!is-a relation} relation. The training dataset's clustering into groups is based on the offset between the vectors of the word pair. The clustering step is expected to uncover hyponymy\index{relation!hyponymy}/hypernymy\index{relation!hypernymy} subrelations. For each cluster, a learned linear projection (in the form of a matrix) represents the hyponymy\index{relation!hyponymy}/hypernymy\index{relation!hypernymy} relation.

Textual patterns\index{pattern} between terms in sentences, encoded by deep-learning methods, can also serve to detect \isa\index{relation!is-a relation} relations. \newcite{Shwartz2016} investigate the effect of combining dependency paths\index{dependency!dependency path} encoded by means of RNNs\index{recurrent neural network (RNN)} with the embeddings\index{embedding!word/entity embedding} of the relation's arguments. All paths between a pair of potential relation arguments participate in producing an averaged representation of the connection between the two arguments. This is assembled from a multi-layered representation of each word on the path. The representation includes the word's lemma and part-of-speech, the dependency label (for the dependency appearing on the considered path), and the direction of the dependency relation\index{dependency!dependency relation}. The method outperforms those based on symbolic distributional models.

\newcite{Le2019} learn the \isa\index{relation!is-a relation} relation from embeddings in a hyperbolic space\index{embedding!hyperbolic space embedding} and from Hearst patterns\index{pattern!Hearst pattern} \cite{hearst92}, a reliable and resilient heuristic for \isa\ in many domains. \citeauthor{Le2019} use Hearst patterns\index{pattern!Hearst pattern} to get \isa\ candidates from a large corpus. From the potential \isa\ instances, they build a ``Hearst graph'', and embed it in a Poincar\'{e} ball. As \newcite{Nickel2018} show, Poincar\'{e} balls are particularly apt for embedding tree structures, necessary in a taxonomy. Constraints on the hyperbolic space enable the detection of erroneous \isa\ instances and the insertion of new \isa\index{relation!is-a relation} links between existing nodes.

\subsubsection{Joint entity and relation extraction}
\vspace{2mm}

Entities and relations\index{relation!relation extraction} can be acquired jointly if one uses their interaction to mutual advantage. Local decisions are made about the text spans which represent entity mentions\index{entity mention span}, argument types\index{argument type} and connections between them. Such decisions constrain named entities\index{named entity} and relations, which can then be learned together \cite{Roth2007}. Deep learning also makes it possible to get and combine such information. We will review methods which can be loosely grouped by how they deal with entity mention identification, and by the mention's varying length:
\begin{packed_item}
\item use some form of span\index{entity mention span} and entity type labelling: sequence labelling\index{sequential labelling} or table-filling with the same style of labels;
\item explore all potential spans\index{entity mention span} (or spans which can be quickly recognized as maybe representing entity mentions), link them in various ways, and predict together the correct spans\index{entity mention span} and the relations between them;
\item process a text fragment, and output relation instances which appear in the text.
\end{packed_item}

\begin{figure}[ht]
 \centering
 \includegraphics[scale=0.45]{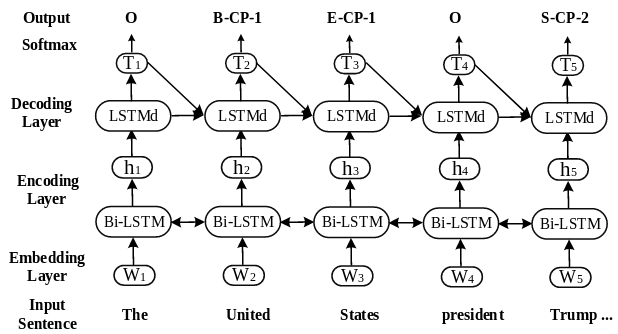}
 \caption{\posscite{Zheng2017} joint entity and relation extraction\index{relation!relation extraction} model.}
 \label{fig:zheng2017model}
\end{figure}

\paragraph{Entity span and type labelling.}%
\newcite{Zheng2017} adopted a tagging scheme, similar to that in \cite{Li2014}, for joint argument identification and relation extraction\index{relation!relation extraction}. Inspired by named entity\index{named entity} tagging, which also must identify sequences of various lengths, \citeauthor{Zheng2017} combined a larger set of span\index{entity mention span} indicators (BIESO: Begin, Inside, End, Single, Other) with the target relations and numerals which indicate the first or the second argument of the relation. Figure \ref{fig:zheng2017model} depicts the model. 

\newcite{Miwa2014} were the first to propose table-filling for relation classification\index{relation!relation classification}. \newcite{Gupta2016} reframe this method as a deep-learning problem. It is the same task: fill a word$\times$word table which corresponds to sentence $s~=~\text{<}w_1, \ldots, w_n\text{>}$. The cells on the diagonal, $(w_i,w_i)$, will be assigned the entity mention span\index{entity mention span} and type labels. A ``regular'' cell $(w_i, w_j), i \ne j$, may be assigned a relation label if $w_i$ and $w_j$ correspond to the head of an entity mention. To perform this labelling, \citeauthor{Gupta2016} choose an ordering of the cells in the table, and process them sequentially\index{sequential labelling} with a context-aware bidirectional RNN.

\newcite{Miwa2016} propose a deep-learning architecture for relation extraction\index{relation!relation extraction}; bidirectional LSTM-RNNs\index{LSTM!bidirectional LSTM} encode the word sequence and the dependency tree\index{parsing!dependency tree}\index{dependency!dependency tree}. \citeauthor{Miwa2016} pretrain the entity identification model and then the relation extraction\index{relation!relation extraction} model with scheduled sampling. Such sampling replaces, with certain probability, predicted entity labels with gold-standard labels. The labels predicted for entity identification guide the selection of candidates for relation classification\index{relation!relation classification}. The entities' heads are detected using the L and U tags, and the candidate pairs for relation classification\index{relation!relation classification} are made from these head words. Compared to \posscite{Li2014} joint model which gives the ACE04 and ACE05 data\index{Automatic Content Extraction (ACE)} to a structured perceptron, \citeauthor{Miwa2016}'s model has better recall and F-score, while the structured perceptron gives higher precision. Compared to CNNs\index{convolutional neural network (CNN)} on the SemEval-2010 Task 8\index{SemEval} data \cite{Santos2015,Xu2015}, there is a higher macro F1-score if bidirectional RNNs\index{recurrent neural network (RNN)} are used with long short-term memory units (BiLSTM-RNNs\index{LSTM!bidirectional LSTM}) to encode the word sequence and the dependency relations\index{dependency!dependency relation}.

\paragraph{Dynamic spans.}%
The basic concept here is also the span\index{entity mention span} of an entity mention. Entity mentions are anchored in those spans, and relations connect two spans. The same entity can be mentioned in different places in a text. Linking different mentions of the same entity brings additional context and information to the detection of the correct span\index{entity mention span} for each mention, and to the selection of the correct relation.

\newcite{Luan2019} build upon the observation that entity identification, relation classification\index{relation!relation classification} and coreference resolution share the common layer of entity mention spans\index{entity mention span}. \citeauthor{Luan2019} develop a multi-task learning framework to tackle the three tasks together. From unstructured input text, their system produces a set of candidate word spans\index{entity mention span}. Each training step identifies the spans most likely to represent entity mentions; those spans serve as nodes in a graph structure. The system then constructs graph edges to represent coreference or semantic relation links, weighted by a confidence score. Span representation is refined by considering contextual information from the predicted relation and from coreference links. \citeauthor{Luan2019} apply their system to four datasets from various domains. \newcite{Wadden2019} build upon this method. They encode spans using contextual language models, and work with task-specific message updates passed over the graph of entity mention spans\index{entity mention span}.

\paragraph{Direct relation extraction.}%
\posscite{Zeng2018} \e{end2end} neural model extracts\index{relation!relation extraction} potentially multiple relation instances directly from a sentence. There is an encoder and a decoder. The encoder transforms the input sentence into a fixed-length semantic vector. The decoder performs three steps to output relation instances based on this semantic vector. It first predicts a relation type. Taking into account the relation type, it determines the source entity, and then copies this entity from the input sentence. Now, given the relation type and the source entity, the decoder determines the target entity, and copies it from the input sentence. \citeauthor{Zeng2018} designed the process in this manner in order to ensure that their system works well if a sentence contains multiple relation instances with overlapping arguments.

\subsection{Learning relations from texts and knowledge graphs}
\label{sec:textgraphDL}

Relation instances in large knowledge repositories\index{knowledge repository}\index{graph!knowledge graph (KG)} often play a role in distant supervision\index{learning!distant supervision} (see Section \ref{sec:distsupnn}), in the learning of relation extraction\index{relation!relation extraction} or classification models\index{relation!relation classification}. It is mutually beneficial to combine evidence from knowledge repositories and unstructured text, and either can help boost the other. This section shows a few examples of successfully combining evidence from unstructured data (either syntactic patterns\index{pattern}, or phrases which connect entities/arguments in a text) with relation instances from knowledge repositories. Such methods have been used to enrich knowledge repositories\index{knowledge repository} with more triples for existing relations or with more relation types, or even to induce a complete relation schema from scratch.

Information from texts and knowledge graphs\index{graph!knowledge graph (KG)} can be merged and then word and relation representations derived jointly, or the two sources of information can be processed separately, and then combined in a final classification step.

\subsubsection{Merging information from texts and knowledge graphs}
\vspace{2mm}

Knowledge graphs\index{graph!knowledge graph (KG)} contain structured information, while unprocessed texts have a linear form. To merge them, texts must also be cast into structures. This can be done in a variety of ways, for example using dependency parsing\index{parsing!dependency parsing}\index{dependency!dependency parsing}, or by extracting specific structured information such as \tripleverb\ triples. The knowledge graphs and the structured textual information can then be merged by mapping nodes and relations, and this bigger structure is processed to produce word/entity and relation/predicate representations which drive relation learning in this hybrid graph. Nodes from KGs and dependency graphs/triples\index{graph!dependency graph}\index{dependency!dependency graph} can be mapped using simple matching, similarity metrics, or more complex entity linking or word sense disambiguation techniques. Relations from the KG\index{graph!knowledge graph (KG)} can also be mapped to predicates or phrases from texts, either before or after the encoding of the merged graph, depending on their induced representations.

\newcite{Lao2012} build such a large graph by combining relation triples from Freebase\index{Freebase} with text processed by a dependency parser\index{parsing!dependency parsing}\index{dependency!dependency parsing}. Pronouns and anaphoric references are clustered\index{cluster, clustering} with their antecedents, and entity mentions are linked to their corresponding nodes from the knowledge repository\index{knowledge repository} by an entity-linking system. To this hybrid graph, \citeauthor{Lao2012} apply the Path Ranking Algorithm (PRA) \cite{Lao2011} which predicts links from paths in knowledge graphs\index{graph!knowledge graph (KG)}. In this case, PRA combines syntactic and semantic cues from the parsed text with relation information to build a model which can predict new relation triples for the knowledge repository\index{knowledge repository}.

In \citeauthor{Lao2012}'s graph, the edges sourced from textual sources are dependency relations\index{dependency!dependency relation}. \newcite{Gardner2013} note that dependency relation names do not contribute semantic information, unlike relations from knowledge repositories\index{knowledge repository}. Instead of dependency graph\index{graph!dependency graph}\index{dependency!dependency graph} representations of texts or text fragments, \citeauthor{Gardner2013} propose to use \tripleverb\ (SVO) triples extracted from texts which parallel \tripleSRT\ triples in knowledge graphs\index{graph!knowledge graph (KG)}; the link in an SVO triple---the predicate---is a lexicalized relation. For connected nodes in the graph built from the knowledge repository\index{knowledge repository}, \citeauthor{Gardner2013} add new edges from SVO triples whose arguments match entities in the graph. There is a difficulty, naturally: adding such predicates directly from large-scale data (600 million SVO triples) would cause an explosion in the number of relation types in the graph, and would not catch equivalent expressions. That is why the lexicalized predicates are replaced with edge labels, which are latent features. These representations are learned by factorizing a $subject\times object$ frequency matrix, built from the SVO data.

The follow-up work gets deeper into the semantic territory, and explores a tighter merging of KGs and texts via the similarity among relations and predicates. \newcite{Gardner2014} work with a graph which combines a KG with SVO triples from texts. They take advantage of the similarity between edge types to allow a random walk to follow edges semantically similar to a given edge type. Nodes obtained from texts and KGs are linked by an \rel{alias} relation, which indicates that the two nodes \e{may} point to the same entity. Edges between subjects and objects extracted from texts are lexicalized predicates, whose vector representation is computed as in \cite{Gardner2013}. To compute the weight corresponding to a path---a sequence of relations $r_1,\ldots,r_n$---for a given $(source, target)$ node pair, at each step $j$ the algorithm can follow either the exact relation type $r_j$ in the path, or another relation type (\ie predicate) close to it in vector space. This allows the score of a ``canonical path'' to combine the score of all (similar) path variations.

Many relation instances may go unnoticed if one restricts links between entities in text to predicates which connect subjects and objects. \newcite{Toutanova2015} treat the lexicalized dependency paths\index{dependency!dependency path}, which they encode using CNNs\index{convolutional neural network (CNN)}, as relations. These semantic representations\index{semantic representation} serve as relation embeddings\index{embedding!relation embedding}\index{relation!relation embedding}; they are combined with evidence from the KG to predict either the target in a $(source, relation, ?)$ query, or the source in $( ?, relation, target)$. The information from the two sources is combined in the model's loss function. One term accounts for the non-negative log-likelihood of the correct entity filler with respect to the graph, computed from a combination of three graph embedding\index{embedding!graph embedding} models. The other term accounts for the non-negative log-likelihood of the correct entity filler with respect to the text; here, the vector representation of the predicate learned by the CNN\index{convolutional neural network (CNN)} replaces the relation representation in the graph embedding\index{embedding!graph embedding} models.

\newcite{Toutanova2016} include all relation paths of bounded length which connect a source and a target node. The paths' contribution is computed as their weighted sum. The contribution of each path is a score which combines the matrix representation of each relation on the path with the weight of the node it connects to.

There is a quantitative and qualitative difference between the predicates obtained via Open IE, and the relations in knowledge repositories\index{knowledge repository}. \newcite{Riedel2013} aim to bridge this gap by deriving a \e{universal schema}\index{universal schema} which combines surface-form predicates retrieved by Open IE with relations already present in knowledge bases\index{knowledge base (KB)}. A very large matrix represents jointly this heterogeneous information: columns correspond to relations from 

\begin{wrapfigure}{r}{0.5\textwidth}
 \centering
 \includegraphics[scale=0.25]{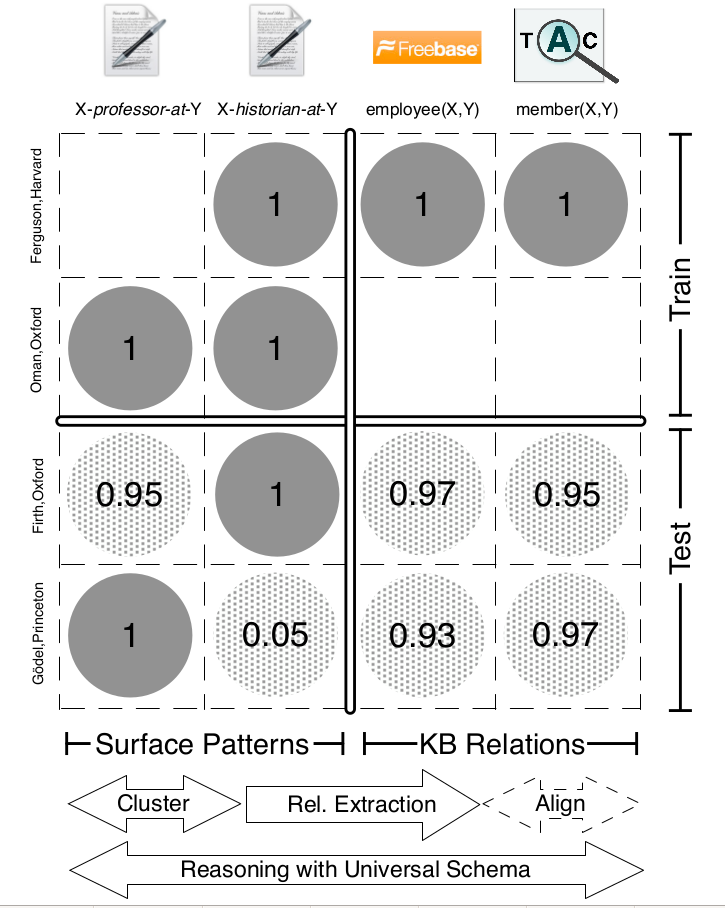}
 \caption{Induction of a universal schema\index{universal schema} from knowledge bases\index{knowledge base (KB)} and texts \cite{Riedel2013}.}
 \label{fig:univSchemaRiedel}
\end{wrapfigure}

\noindent knowledge repositories\index{knowledge repository} and predicates found in texts; rows correspond to entity/word pairs. A cell is marked if the corresponding entity pair appears in the given relation or context---see Figure \ref{fig:univSchemaRiedel}. Matrix factorization\index{matrix factorization} induces representations of the entity pairs and the relations/syntactic patterns\index{pattern}, in the manner explained in Section \ref{sec:wefromgraphs}. From such representations, one can determine associations between syntactic patterns and semantic relations, and map these lexical expressions onto the ``canonical'' relation form. The representations can also help cluster\index{cluster, clustering} syntactic patterns\index{pattern} to indicate new (unnamed) relations, not yet included in the knowledge repository\index{knowledge repository}.

\begin{figure}[ht]
 \centering
 \includegraphics[clip=true, trim=1cm 0cm 0cm 0cm, scale=0.33]{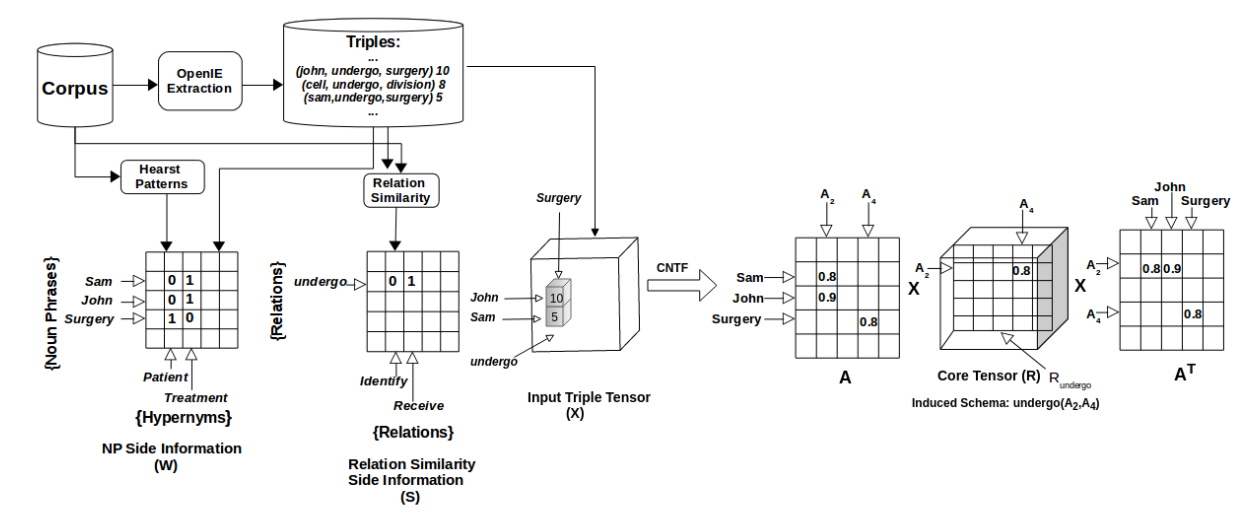}
 \caption{Relation schema induction from unstructured texts, using Hearst patterns\index{pattern!Hearst pattern} and Open IE triples \cite{Nimishakavi2016}.}
 \label{fig:nimi}
\end{figure}

~ \newcite{Nimishakavi2016} derive a universal schema\index{universal schema} and a knowledge repository\index{knowledge repository} from unstructured text in a specific domain; they do it without the benefit of a ``seed'' knowledge repository. Instead of such prior knowledge, \citeauthor{Nimishakavi2016} gather two types of ``side information'' to help structure, and make canonical, candidate \tripleverb\ triples extracted by Open IE methods. The side information consist of hyponym\index{relation!hyponymy}/hypernym\index{relation!hypernymy} candidates extracted using Hearst patterns\index{pattern!Hearst pattern}, and relation similarity (as similarities between verbs in the Open IE triples). The extracted triples are represented in a tensor\index{tensor}, factorized together with the side information to induce the relation schema. Figure \ref{fig:nimi} illustrates the method.

\newcite{Riedel2013} developed methods based on scoring functions\index{scoring function} for \tripleSRT\ triples, which combine partial scores on various pairs of the three elements. This limits the applicability of the methods to already seen $source$-$target$, $source$-$relation$ or $relation$-$target$ pairs. \newcite{Verga2017} address this limitation with a representation for a pair of entities based on the textual patterns\index{pattern} in which they appear. \newcite{Verga2016} extend the application of a universal schema\index{universal schema} to multilingual data: they leverage common representations for shared entities, and match the textual patterns\index{pattern} in the representation with relations in the knowledge repository\index{knowledge repository}.

\newcite{Zhang2019} encode every entity; their model combines these representations with vector representations of the target (KB) relations, and with attention\index{attention} methods for relation prediction. The representation of a source or target entity is based on its neighbourhood in the knowledge graph\index{graph!knowledge graph (KG)}, and on its co-occurrences in the triples extracted from texts. \citeauthor{Zhang2019}'s method is applied to Freebase\index{Freebase} and to a subset of Freebase\index{Freebase} with film-related relations, as well as to triples extracted from IMDB \cite{Lockard2019} and ReVerb extractions from ClueWeb whose subject is linked to Freebase\index{Freebase} \cite{Lin2012}.

\subsubsection{Knowledge graphs and texts as separate information sources}
\vspace{2mm}

The merging of information from texts and knowledge graphs\index{graph!knowledge graph (KG)} aims to build a larger graph which can be processed with methods similar to those developed for processing KGs: link prediction\index{link prediction} using paths, matrix factorization\index{matrix factorization}, and so on. Without casting texts in structured forms, they can provide additional information about the nodes or the relations in the graph, or an additional signal for relation learning in KGs\index{graph!knowledge graph (KG)}.

\newcite{Weston2013} encode information from the knowledge graph\index{graph!knowledge graph (KG)} and the textual context separately, and use them together for relation extraction\index{relation!relation extraction}. The KG is encoded with \posscite{Bordes2013} translation model, and TransE's scoring function\index{scoring function} provides one part of the information. Information from texts is encoded by a function which computes a similarity measure between a relation mention and a relation embedding\index{embedding!relation embedding}\index{relation!relation embedding}. A $(source, target)$ pair is assigned the relation \rel{rel} with the highest score. This score combines the $(source, rel, target)$ triple's KG score (TransE) and a text-based score; the latter is the cumulative score for the similarity of \rel{rel}'s embedding to every mention of the arguments (\ie every sentence which contains $source$ and $target$).

\newcite{Xie2016} combine textual evidence and knowledge base\index{knowledge base (KB)} relations by associating textual descriptions with entities in the KB. They encode relational triples with \citeauthor{Bordes2013}'s TransE while inducing a representation of the entities which can be useful in predicting their textual descriptions. For the purpose of learning, these textual descriptions (included in Freebase\index{Freebase}) are encoded with two formalisms: continuous bag-of-words and convolutional neural networks\index{convolutional neural network (CNN)}. The method produces entity representations which capture both the relational information and their descriptions, and that affects the encoding of the TransE-derived relations. The representations give link prediction\index{link prediction} results better than any of the subsumed formalisms.

\newcite{Fan2016} follow on \citeauthor{Xie2016}'s work. They reduce the number of parameters of the model, and cast it into a probability framework. The improved model measures the probability of each relational triple, and maximizes the log-likelihood of the observed knowledge to learn simultaneously the contextualized embeddings\index{embedding!contextualized embedding} of entities, relations and words in descriptions. \newcite{Zhang2019longtail} apply knowledge graphs\index{graph!knowledge graph (KG)} (Freebase\index{Freebase}), texts and pretrained word embeddings\index{embedding!word/entity embedding} to the problem of long-tail relations. The encoding of sentences which include specific relations helps supplement the information about low-frequency entity pairs. Hierarchical information for Freebase\index{Freebase} relations (as revealed by their names) goes into a graph convolution network\index{graph convolution network (GCN)} to induce similar representations for similar relations. This helps derive informative representations for low-frequency relations.

Natural language understanding---in particular reading comprehension---is one of the high-end tasks to which semantic relations can, or indeed should, contribute. Reading comprehension systems are commonly tested on question answering\index{question answering} (QA) \cite{Light2000}. \newcite{Levy2017} show how to model the relation extraction\index{relation!relation extraction}/classification\index{relation!relation classification} task as a QA\index{question answering} task. They adopt a slot-filling framework---look for the entity to complete a triple $(e_1, rel, ?)$---where the relation ranges over all relation types plus \rel{no relation}. To frame this as a QA\index{question answering} problem, \citeauthor{Levy2017} first \e{querify} the incomplete relation triple: transform it into a query/question using crowd-sourced templates. To extract a relation instance\index{relation!relation extraction}, a bi-directional attention\index{attention} flow network \cite{Seo2017} is given as input a sentence with a potential relation instance, together with a query which pairs up a source entity with a relation type. The system outputs a span\index{entity mention span} which corresponds to the target entity, or signals the absence of an answer.

It is an important characteristic of this method that it can generalize to relation types it was not trained on. That is because the model does not learn to associate specific relation types with a given context, but learns to focus on a sentence segment most relevant to the query. There are further developments in this line of research: incorporate relation extraction\index{relation!relation extraction} as a QA\index{question answering} task in a multi-task setting \cite{McCann2018}; and apply multi-turn QA\index{question answering} \cite{Li2019QA} which allows sequential discovery of relations, potentially using previous results to answer the current question.

\subsection{N-ary and cross-sentence relations}
\label{sec:naryDL}

Most research focuses on binary semantic relations. Even so, n-ary relations are often necessary to acquire sufficient knowledge, especially in specialized domains such as chemistry or medicine. Such relations may also be expressed over a number of sentences, and that makes their extraction\index{relation!relation extraction} even more difficult. Consider this fragment from the biomedical literature \cite{Heuckman:ALK}:\footnote{The summary results in the paper include the following statement: ``An independent resistance screen in \e{ALK}-mutant neuroblastoma cells yielded the same L1198P resistance mutation but defined two additional mutations conferring resistance to TAE684 but not to PF02341066.'' That is to say, the entities \e{ALK} and \e{PF02341066} are related.}
\begin{quote}
\e{%
We next expressed \ul{ALK}$^{F1174L}$, \ul{ALK}$^{F1174L/L1198P}$, \ul{ALK}$^{F1174L/G1123S}$, and \ul{ALK}$^{F1174L/G1123D}$ in the original SH-SY5Y cell line. \\~\\
}
\lbrack\ldots 15 sentences in 3 paragraphs\ldots\rbrack \\~\\
\e{%
The 2 mutations that were only found in the neuroblastoma resistance screen (G1123S/D) are located in the glycine-rich loop, which is known to be crucial for ATP and ligand building and are the first mutations described that induce resistance to TAE684, but not to \ul{PF02341066}.
}
\end{quote}

Interestingly, n-ary relations were a target at the first Message Understanding Conference.\footnote{\url{ir.nist.gov/muc}} The task was to determine the attributes of an event (who, where, when, and so on), but each of the n-ary relations was split into binary subrelations, and each of those was dealt with via binary relation extraction\index{relation!relation extraction}/classification\index{relation!relation classification}.

\newcite{Chen2019} similarly treat n-ary relation extraction\index{relation!relation extraction} as a collection of binary relation extraction\index{relation!relation extraction} subtasks. They also allow an explicit adjustment of the context window size up to two sentences. Working in a narrow domain (clinical corpus on breast cancer treatments) with limited data, \citeauthor{Chen2019} find that the results improve when the text is modelled in terms of phrases and recognized concepts, and enriched with word embeddings\index{embedding!word/entity embedding} and synonyms\index{relation!synonymy}. A support vector machine\index{Support Vector Machine (SVM)} gives better results than a feed-forward neural network with two fully connected layers.

\posscite{Akimoto2019} system for n-ary relation extraction\index{relation!relation extraction} combines universal schemas\index{universal schema} and the decomposition of n-ary relations into unary and binary relations. Representations for unary and binary relations found in a knowledge base\index{knowledge base (KB)} and in text are learned from the training data. The learning of the model for n-ary relations relies on optimizing a score which aggregates the lower-arity relation scores.

\newcite{Quirk2017}, \newcite{Peng2017} and \newcite{Wang2018} tackle the cross-sentence relation extraction\index{relation!relation extraction} task by taking into account a context larger than a sentence. They all combine inter-sentential relations (grammatical dependencies\index{dependency} and word sequence information) with discourse relations\index{relation!discourse relation} and sentence-level sequence information. \citeauthor{Peng2017} give this document-graph structure as input to a BiLSTM\index{LSTM!bidirectional LSTM}, as illustrated in Figure \ref{fig:docgraph}. The forward pass takes the word sequence information and forward-looking dependencies\index{dependency}; the backward pass takes the reversed word sequence information and the backward-looking dependencies\index{dependency}. The word representations derived by this formalism become the input to a relation classification\index{relation!relation classification} step. \newcite{Peng2017} classify every entity mention pair in their document graph.

This form of relation extraction\index{relation!relation extraction} does not scale well beyond one document because of the combinatorial explosion of entity mention combinations at such a high level. \posscite{Jia2019} remedy is an entity-centric model: mentions are first mapped onto entities, and entity combinations are explored.

\newcite{Christopoulou2019} build a document graph as well, but rely on different kinds of information. To obviate the need for grammatical properties, they use occurrence or co-occurrence to connect nodes which correspond to mentions, entities and sentences: mentions are connected to the sentences in which they occur, to the entities they correspond to, and to other mentions they cooccur with in a sentence; entities are connected to sentences which contain one of their mentions. \citeauthor{Christopoulou2019} construct node descriptions from word descriptions. A mention and a sentence are represented by an average of the representations of their words, while an entity's representation averages the representations of its mentions. The aim is to build and represent edges between pairs of entities. The various paths between two entities are aggregated iteratively into edge representations. These representations are then used to classify every edge into a relation type.

\posscite{Verga2018} model predicts mentions and relations at document level. Mention identification is simulated with an attention\index{attention} mechanism over tokens whose representation combines the actual token embedding\index{embedding!word/entity embedding} and the positional embedding. This input is passed through several layers of multi-head attention\index{attention}---to attend to different types of relevant information---and through convolution components. From the output of this process, \citeauthor{Verga2018} build two position-specific representations, for the head and the tail (the source and the target) of a relation. These representations yield a pair-wise relation affinity tensor\index{tensor}, which drives the final relation prediction.

The work discussed thus far relies on document-level information---entity mentions, and intra- and inter-sentence relations---processed together. \newcite{Singh2019} predict single relations which may cross sentence boundaries. They rely on additional context tokens which mediate the targeted relation\index{relation!targeted relation}. The motivation is to deal with entity mention pairs distant in the text. Finding intermediary tokens in a relation with each of the targeted mentions can help address the distance problem, and give clues about the interaction of the original pair. \citeauthor{Singh2019} account for these \e{second-order relations} in a transformer-based model. The model initially scores the first-order relations to intermediary tokens. Next, it scores the second-order relations by aggregating the scores for all first-order relations which mediate the targeted second-order relations.

\subsection{Unsupervised relation extraction}
\label{sec:unsupDL}

There are several ways of tackling unsupervised\index{learning!unsupervised learning} relation extraction\index{relation!relation extraction} which predate the deep-learning period. They rely on semantic similarity to group extracted tuples. The similarity is calculated between the arguments of different relation instances or between the patterns\index{pattern} which those instances display. Similarities can be used directly to find the closest relation instance match, or to cluster\index{cluster, clustering} similar instances.

A good representation of a sentence with an instance of a relation $r$ should be close to the representations of other sentences with other instances of $r$. This assumption can lead to ``implicit'' clusters\index{cluster, clustering}; \newcite{Marcheggiani2016} rely on it for their variational autoencoder model. The encoder builds a semantic representation\index{semantic representation} for a sentence based on a feature-rich representation. The expectation is that the representation so built will approximate the representation of a relation triple. The decoder can then reconstruct one of the arguments of the relation. The two components are trained together. In the encoding step, the argument to be reconstructed is obscured. 

Information from a knowledge base\index{knowledge base (KB)} added to this model introduces similarity constraints between relation tuples. \newcite{Liang2019} learn to discover instances of previously unseen relations. They expand \citeauthor{Marcheggiani2016}'s model using the similarity between two entity pairs $x_1 = (x_{11},x_{12})$ and $x_2 = (x_{21},x_{22})$ as the cosine of the angle between the translation vectors connecting the entities in each pair:
\begin{align}
sim(x_1,x_2) = cos(\embv_{11}-\embv_{12},\embv_{21}-\embv_{22})
\end{align}
$\embv_{ij}$ is the knowledge base\index{knowledge base (KB)} embedding of entity\index{embedding!word/entity embedding} $x_{ij}$. 

\citeauthor{Liang2019} compute two variations. One of them represents the \e{must-link} confidence score $s^+(x_1,x_2) = [sim(x_1,x_2)]^+_{\gamma^+}$, the other the \e{must-not-link} confidence score $s^-(x_1,x_2) = [sim(x_1,x_2)]^-_{\gamma^-}$, where the thresholds $\gamma^+,\gamma^- \in [0,1]$ limit the two scores. ($[x]^+_{\gamma^+}= \mbox{ if } x > \gamma^+ \mbox{ then } x \mbox{ else } 0$; $[x]^-_{\gamma^-}= \mbox{ if } x < -\gamma^- \mbox{ then } x \mbox{ else } 0$.) These scores, together with the scores which compare the corresponding sentence representations, help determine if the two sentences containing these relation tuples should be in the same cluster\index{cluster, clustering}, \ie should represent the same relation. The sentence representations are derived by a system built upon \citeauthor{Marcheggiani2016}'s variational autoencoder.

\posscite{Papanikolaou2019} method generates its own training data for targeted relations\index{relation!targeted relation} on a pre-specified list. They first extract \tripleverb\ triples, and use pre-trained embeddings to map the verbs onto the given set of target relations. The system drops verbs whose similarity to any given relation falls below a threshold. The triples with the accepted verbs are mapped onto texts, as in distant supervision\index{learning!distant supervision}. The sentences which contain the automatically annotated relation instances then help fine-tune a BERT\index{Bidirectional Encoder Representations from Transformers (BERT)} model for relation classification\index{relation!relation classification}.

\subsection{Lifelong Learning}
\label{sec:lifelongDL}

Deep learning requires very large training data to build accurate models. The model consists of the network architecture and its parameters---weights in its various units---whose best values are determined during training. Training such a model costs a great deal of computing time and power. A deep-learning system which aims to continue learning faces a dilemma. It can keep retraining on ever-growing datasets, or be doomed to forget much of what it has learned if it gets none, or only a subset, of the old data together with newer instances for training a new model. That is because even a small change in the learned parameters (when the model is updated on new data) may cause unpredictable behaviour on the older data.

\newcite{Wang2019} suggest a two-part solution. Inspired by previous research on handwriting and object recognition, they propose a new strategy: maintain a ``training memory'', and select instances from previously used data to add to a new training set, and so avoid forgetting the older data. They call their method \e{episodic memory replay} (EMR). The ``memory'' $\mathcal{M}$ consists of a number of training examples selected after each training session. When training on a new dataset, EMR adds a sample of instances from $\mathcal{M}$ to the current training data, so the model can retain the knowledge of previous data. The second part of the solution arises from the observation that a good model should not distort excessively the embedding space\index{embedding!embedding space} of the model's parameters when it gets additional training data. For the task of relation extraction\index{relation!relation extraction} in particular, \citeauthor{Wang2019} use the sentence embeddings\index{embedding!sentence embedding} derived by the neural model in previous sessions as anchor points, and constrain the system to only minimally distort these anchor points with the processing of new data.

%% %% section %% %%
\section{Summary}
\label{sec:deepSummary}

This chapter has presented an overview of the recognition and classification of semantic relations in the deep-learning paradigm. The methods developed for the traditional learning of semantic relations can be directly mapped onto this new formalism but the power of deep learning is best unleashed when we can take advantage of its specific characteristics:
\begin{packed_item}
\item low-dimensional representation of word meaning based on various types of knowledge;
\item representation derived simultaneously for arguments and relations;
\item the leveraging of multiple information sources;
\item the availability of formalisms which encode variable-length sequences and find patterns\index{pattern} in them;
\item the encoding of graph structures (syntactic or semantic) together with a variety of additional attributes.
\end{packed_item}

Deep learning requires, among other things, large amounts of training data. Some such data can be bootstrapped from existing knowledge repositories\index{knowledge repository} by distant supervision\index{learning!distant supervision}. The adoption of deep learning has led to innovative ways of cleaning the automatically annotated data. Many interesting methods have been developed to tackle noise in automatically generated data: people have applied adversarial learning\index{learning!adversarial learning}, reinforcement learning\index{learning!reinforcement learning} and other fun formalisms. The new technology has opened a vast space of exploration. We have presented some of the main trends, but there are new directions to be found, and space in between. 

\clearpage

%\input{ch06}
%\blankpage

%\begin{thebibliography}{xx}
\addcontentsline{toc}{chapter}{\numberline{}{Bibliography}}

\markboth{\MakeUppercase{Bibliography}}{}

\bibliographystyle{plainnat}
\bibliography{SRBN2}

%\end{thebibliography}
\clearpage % bibliography
\newpage

\backmatter % back matter
%\blankpage % before the bios; maybe unnecessary
%\input{bios} % authors' short bios
%\input{SRBN2.ind} % if the author has Index

%\blankpage % before the index; maybe unnecessary
\printindex
\clearpage % after the index
\end{document}